\documentclass{article}

\PassOptionsToPackage{round, comma, sort&compress, authoryear}{natbib}
\usepackage[preprint]{neurips_2025}
\usepackage[T1]{fontenc}
\usepackage[utf8]{inputenc}
\usepackage{microtype}
\usepackage{textcomp}
\PassOptionsToPackage{hyphens}{url}
\usepackage{url}
\usepackage{amsmath}
\usepackage{amssymb}

\usepackage{booktabs}
\usepackage{array}
\usepackage{longtable}
\usepackage{calc}

\usepackage{enumitem}

\usepackage[font=small,labelfont=bf]{caption}

\usepackage{graphicx}
\graphicspath{{figures/}}

\usepackage{xcolor}

\usepackage[colorlinks=true,
            linkcolor=blue!50!black,
            citecolor=blue!50!black,
            urlcolor=blue!50!black]{hyperref}
\usepackage[capitalize, noabbrev]{cleveref}

\title{How Language Models Organize and Structure Moral Knowledge}

\author{%
  Orion Reblitz-Richardson\thanks{%
    Distiller Labs. Correspondence to Distiller Labs
    \textless\texttt{orion@orionr.com}\textgreater.}
}

\date{August 2026}

\providecommand{\tightlist}{%
  \setlength{\itemsep}{0pt}\setlength{\parskip}{0pt}}

\usepackage{fancyvrb}
\DefineVerbatimEnvironment{Highlighting}{Verbatim}{commandchars=\\\{\}}

\begin{document}

\maketitle

\begin{abstract}
How do large language models (LLMs) organize moral knowledge? Models
detect moral content broadly, but detection is a low bar. We ask
whether they go further, distinguishing moral foundations from one
another and organizing the relationships between them geometrically.

We train six independent linear probes on open-weight language models,
one per Moral Foundations Theory (MFT) category (\emph{care/harm},
\emph{fair/cheat}, \emph{lib/oppress}, \emph{loy/betray},
\emph{auth/subv}, \emph{sanc/degrade}), and examine how the resulting
directions relate to each other in representation space. We find the
directions neither collapse into a single moral detector nor isolate
from one another. Rather, they span a near-maximal number of
independent dimensions while sharing a positive common component. The
shared component is \textbf{the signature of integration}, and it is
moral-specific relative to a matched non-moral concept battery built
identically (mean pairwise cosine 0.26 vs.\ 0.013).

The geometry is consistent across architectures and scale and reaches
its integration regime early in pre-training, well before probe
accuracy saturates. The structure the model discovers shows
\textbf{no evidence} of the individualizing/binding distinction
predicted by Moral Foundations Theory (an underpowered test: only 10 distinct splits exist, so it cannot reject at the 0.05 level) but rather reflects corpus statistics.
Extending to moral dilemmas, each dilemma direction partially composes
from its component foundations, at $2.7\times$ a mismatched-pair
baseline, while the majority of its variance encodes conflict-specific
structure. The model represents moral tension itself, not a
pre-resolved judgment.
\end{abstract}

\section{Introduction}\label{introduction}

How do language models represent morality? Prior work in this
series established that models encode moral content \emph{broadly}:
probing accuracy saturates early during pre-training and spans
nearly all layers \citep{reblitzrichardson2026fragility}. Fragility
testing revealed that this encoding grows more robust throughout
training even after accuracy plateaus. Extension to
mixture-of-experts (MoE) models showed that moral signal is
uniform across experts, with no expert specialization, but that a
74\(\times\) output scale gap produces structural fragility
\citep{reblitzrichardson2026dilution}.

Both papers treated moral encoding as a single binary feature:
moral vs.~neutral. A model that merely detects ``this text involves
morality'' has cleared a low bar. Genuine moral understanding
requires \emph{structured} representations: the ability to distinguish
care from fairness, loyalty from authority, and to encode the
relationships between them. The transition from moral \emph{detection}
to moral \emph{understanding} is the subject of this paper.

We operationalize this transition through the geometry of
foundation-specific probe directions. Where prior work trained one
probe to separate moral from neutral content, we train six: one for
each Moral Foundations Theory (MFT) foundation
\citep{haidt2012righteous, graham2013mft}. The learned probe weight
vectors define \emph{directions} in the model's representation space.
The angular relationships between these directions reveal whether
the model has developed structured moral representations.

Three geometric signatures correspond to three qualitatively
different modes of moral representation:

\begin{enumerate}
\def\labelenumi{\arabic{enumi}.}
\item
  \textbf{Collapse.} All foundation directions converge toward a single
  ``moral salience'' direction. The model detects moral relevance
  but does not distinguish frameworks.\footnote{We use \emph{framework}
    interchangeably with the Moral Foundations Theory term \emph{foundation}
    throughout (and in the title). We do not intend the moral-psychology
    sense in which \emph{framework} denotes a theory-level position such as
    deontology or utilitarianism. We retain \emph{framework} in part because
    \emph{foundation} is itself overloaded in the language-model setting,
    where a ``foundation model'' is a base, pre-fine-tuning model,
    exactly the class of models we probe.} This is detection without
  structure.
\item
  \textbf{Isolation.} Foundation directions are orthogonal with no
  relational structure. The model has separate moral ``slots'' but
  no representation of how frameworks relate. This is structure
  without coherence.
\item
  \textbf{Integration.} Foundation directions are separated but
  non-orthogonal, with inter-framework geometry reflecting known
  relationships from moral psychology. This is the precondition
  for moral reasoning.
\end{enumerate}

Applied to OLMo-2 1B and OLMoE-1B-7B (with a dense OLMo-2 7B as a
scale check), we report three findings:

\textbf{Finding 1: Moral foundations are represented with integration
geometry, but we find no evidence of the structure moral psychology
predicts.}
Foundation directions are distinct: mean pairwise cosine similarity is
\(\approx 0.22\)--\(0.27\) across layers, positive everywhere and far from
collapse. This positive shared component is the integration signature.
Against a matched non-moral concept battery built identically to the
foundations, it is \({\sim}20\times\) larger (0.26 vs.~0.013; paired
\(\Delta = 0.223\), CI \([0.202, 0.244]\), excluding 0; \Cref{framework-geometry-integration-not-collapse}), so the shared
component is moral-specific relative to a matched non-moral battery
rather than a generic content-vs-neutral axis (whether it is
specifically moral rather than generic affective salience is the one
residual control we flag; \S5.6). The six directions span 5 effective
dimensions at
every layer, which rules out collapse but, at the ceiling for six
mean-centered directions, does not by itself separate integration from
isolation. Hierarchical clustering does not recover the MFT
individualizing/binding split at either the dense 1B or 7B, though the
group-structure test is underpowered (smallest achievable \(p = 0.10\),
so it cannot reject at \(\alpha = 0.05\) and a small effect is not
excluded); the most consistent structure the
model forms is a care--sanctity pairing that crosses MFT groups.

\textbf{Finding 2: The geometry of moral dilemmas is partially
compositional.} Probes for 15 two-foundation dilemmas reach 94.2\%
mean accuracy, and each dilemma direction is partly explained by the
2D subspace of its two component foundation directions. Dilemma pairs
that share a component foundation are closer in representation space
than those that do not (mean cosine 0.273 vs.~0.196, permutation
\(p = 0.0001\)), and the MoE architecture preserves this compositional
structure.

\textbf{Finding 3: Output dilution degrades every foundation uniformly,
not selectively.} Extending the fragility protocol to per-foundation
probes shows a uniform cross-architecture effect rather than a
per-foundation one: once accuracy is averaged over multiple noise
seeds, every foundation is more fragile in MoE than in dense (a
\({\sim}2.3\times\) gap), and no single foundation is reliably most or
least robust within an architecture. An apparent complexity--fragility
ordering does not survive scale normalization (\S4.11). Output dilution
suppresses moral encoding across the board.

This paper introduces probe-direction geometry, a method for
measuring structured moral representations that bridges binary moral
probing and the richer structure posited by moral psychology. With
it, we give the first geometric characterization of foundation and
dilemma representations in base language models, and show that this
geometry is comparable across dense and MoE architectures and stable
from 1B to 7B. We also report the first per-foundation fragility
comparison across architectures, which finds that output dilution
degrades moral encoding uniformly rather than selectively.

\section{Related Work}\label{related-work}

\subsection{Probing for linguistic and semantic structure}\label{probing-for-linguistic-and-semantic-structure}

Linear probing \citep{alain2017probes} has become a standard tool
for reading off what information is encoded in neural network
representations. \citet{conneau2018probing} probed sentence
embeddings for syntactic properties; subsequent work probed for
part-of-speech, dependency relations, coreference, and world
knowledge. The linear representation hypothesis
\citep{park2024linear} formalizes the claim that concepts are
encoded as directions in representation space, precisely the
assumption underlying our use of probe weight vectors as geometric
objects.

Two limitations of the probing literature motivate our approach.
First, most probing studies report only \emph{accuracy}, discarding the
learned probe parameters. We show that the probe weight vector
(the normal to the classification hyperplane) carries geometric
information about how concepts relate to each other. Second,
probing studies typically train one probe per concept, treating
concepts as independent. Our multi-probe geometric analysis
recovers inter-concept structure from independently trained probes.

\subsection{Geometry of concept representations}\label{geometry-of-concept-representations}

\citet{bolukbasi2016gender} demonstrated that gender bias in word
embeddings manifests as a geometric subspace, and that debiasing
amounts to projecting out a direction. \citet{arditi2024refusal}
showed that refusal behavior in instruction-tuned LLMs is mediated
by a single direction in activation space. These results establish
the precedent that high-level behavioral properties can be localized
to specific directions.

Our work extends this line from single directions to \emph{sets of
related directions}. Where prior work asked ``where is concept \(X\)?''
we ask ``what is the geometric relationship between concepts
\(X_1, \ldots, X_k\)?'' The cosine similarity matrix between
foundation probe directions is a form of representational similarity
analysis \citep[RSA;][]{kriegeskorte2008rsa} applied not to stimulus
response patterns but to the probes that decode them.

\subsection{Moral psychology and Moral Foundations Theory}\label{moral-psychology-and-moral-foundations-theory}

Moral Foundations Theory \citep[MFT;][]{haidt2012righteous,
graham2013mft} posits that human moral judgment draws on (at least)
five or six innate foundations: care/harm, fairness/cheating,
loyalty/betrayal, authority/subversion, sanctity/degradation, and
(later added) liberty/oppression. MFT predicts a structural
distinction between \emph{individualizing} foundations (care, fairness,
liberty), which protect individuals from harm, and \emph{binding}
foundations (loyalty, authority, sanctity), which bind individuals
into groups. This distinction has been validated in cross-cultural
surveys and predicts political orientation
\citep{graham2013mft}.

Alternative taxonomies exist. \citet{curry2019cooperation} propose
morality-as-cooperation, identifying seven moral domains grounded
in evolutionary game theory. Our use of MFT is pragmatic: the
six-foundation taxonomy gives a tractable set of directions for
geometric analysis, and the individualizing/binding prediction
gives a testable structural hypothesis.

\subsection{Moral reasoning in language models}\label{moral-reasoning-in-language-models}

\citet{hendrycks2021ethics} introduced the ETHICS benchmark for
evaluating moral reasoning in language models across justice,
deontology, virtue, utilitarianism, and commonsense dimensions.
Most work in this area evaluates models \emph{behaviorally},
measuring what outputs models produce in response to moral
scenarios. Our approach is \emph{representational}: we probe the
internal geometry of moral encoding, asking not whether the model
produces the right moral judgment but whether it has developed
structured representations of moral distinctions.

\subsection{Companion papers}\label{companion-papers}

This paper is the third in a series. Paper 1
\citep{reblitzrichardson2026fragility} established that moral
content is decodable by linear probes from the earliest layer
of OLMo-2, that probing accuracy saturates early during
pre-training, and that fragility testing (injecting Gaussian
noise into activations) resolves the continued development of
moral encoding after accuracy plateaus. Paper 2
\citep{reblitzrichardson2026dilution} extended the analysis to
MoE models (OLMoE-1B-7B), finding uniform moral encoding across
experts but a 74\(\times\) output scale gap that produces structural
fragility.

The present paper moves from \emph{binary} moral encoding (moral
vs.~neutral) to \emph{structured} moral encoding (framework-specific
directions and their inter-framework geometry). It reuses the
probing and fragility protocols from Papers 1 and 2, applying
them at the per-foundation level.

\section{Methodology}\label{methodology}

We train six foundation-specific linear probes at each transformer
layer, extract the learned weight vectors as geometric directions in
representation space, and analyze the angular relationships between
these directions. The methodology decomposes into five components:
foundation-specific probing (\Cref{foundation-specific-probing}), geometric analysis of probe
directions (\Cref{geometric-analysis}), bootstrap direction stability assessment (\Cref{bootstrap-direction-stability}),
framework-specific fragility testing (\Cref{framework-specific-fragility}), and the probing
dataset (\Cref{probing-dataset}). All experiments run on a single MacBook Pro M4 Pro
(24 GB unified memory, MPS backend).

\subsection{Models and comparison design}\label{models-and-comparison-design}

We evaluate three base (non-instruct) models from the same lab
(Ai2). The primary comparison is between two architectures matched in
layer count (16), dense vs.~mixture-of-experts; a third, larger dense
model serves as a scale control (\S4.14):

\begin{itemize}
\item
  \textbf{OLMo-2 1B} (\path|allenai/OLMo-2-0425-1B|): dense transformer,
  1.5B parameters, 2048 hidden dimension, 16 layers \citep{olmo2_2025}.
  Ai2 publishes 37 early-training checkpoints at 1K-step intervals
  (steps 0--36K), enabling trajectory analysis.
\item
  \textbf{OLMoE-1B-7B} (\path|allenai/OLMoE-1B-7B-0924|): mixture-of-experts,
  6.9B total parameters, 1.3B active, 64 experts per layer, top-8
  routing, 2048 hidden dimension, 16 layers \citep{muennighoff2024olmoe}.
\item
  \textbf{OLMo-2 7B} (\path|allenai/OLMo-2-1124-7B|): dense transformer, 7.3B
  parameters, 4096 hidden dimension, 32 layers \citep{olmo2_2025}. Used
  in \S4.14 to test whether the geometry findings persist with scale.
\end{itemize}

All three models use the same tokenizer and comparable training corpora.
The comparison tests whether framework geometry is an artifact of
dense connectivity or a general property of transformer-based
language modeling. \citet{reblitzrichardson2026dilution} established
that moral encoding in OLMoE is uniform across experts with a
74\(\times\) output scale gap relative to OLMo-2; the present work
asks whether this output dilution affects the \emph{structure} of moral
representations, not just their \emph{scale}.

\subsection{Foundation-specific probing}\label{foundation-specific-probing}

For each of the six Moral Foundations Theory (MFT) foundations
(care/harm, fairness/cheating, loyalty/betrayal, authority/subversion,
sanctity/degradation, and liberty/oppression;
\citep{haidt2012righteous, graham2013mft}), we train a binary
linear probe at each of 16 transformer layers.

\textbf{Probe architecture.} Each probe is \texttt{nn.Linear(2048,\ 1)} trained
with BCE loss and Adam (lr = \(10^{-2}\)) for 50 epochs. This
matches the probe specification from
\citet{reblitzrichardson2026fragility} and
\citet{reblitzrichardson2026dilution}, enabling direct comparison
with their binary moral/neutral probes.

\textbf{Activation collection.} For each text, we run a single forward
pass capturing all 16 layers simultaneously. At each layer, we
mean-pool across the sequence dimension to obtain a single
\(\mathbb{R}^{2048}\) vector per text. Positive examples are the
foundation-tagged moral sentences; negative examples are their
matched neutral counterparts.

\textbf{Direction extraction.} After training, we extract the weight
vector \(\mathbf{w} \in \mathbb{R}^{2048}\) from each probe and
normalize to unit length: \(\hat{\mathbf{w}} = \mathbf{w} /
\|\mathbf{w}\|\). This unit vector is the normal to the
classification hyperplane, the \emph{direction} in representation
space that maximally separates the foundation's moral content from
neutral content. We call \(\hat{\mathbf{w}}\) the foundation's
\textbf{probe direction} at a given layer.

This yields \(6 \times 16 = 96\) unit-norm probe direction vectors
per model.

\subsection{Geometric analysis}\label{geometric-analysis}

The paper's core contribution is analyzing the angular relationships
between the six foundation probe directions at each layer.

\textbf{Cosine similarity matrices.} At each layer, we compute the
\(6 \times 6\) pairwise cosine similarity matrix of the foundation
probe directions. Because the directions are unit-normalized,
\(\cos(\hat{\mathbf{w}}_i, \hat{\mathbf{w}}_j) =
\hat{\mathbf{w}}_i \cdot \hat{\mathbf{w}}_j\).

\textbf{Geometric signatures.} Three qualitative modes of moral
representation correspond to distinct cosine similarity patterns:

\begin{enumerate}
\def\labelenumi{\arabic{enumi}.}
\item
  \emph{Collapse (averaging).} All foundation directions converge
  toward a single ``moral salience'' direction. Mean pairwise
  cosine similarity \(\to 1\).
\item
  \emph{Isolation.} Foundation directions are orthogonal with no
  relational structure. Mean pairwise cosine similarity \(\to 0\).
\item
  \emph{Integration.} Foundation directions are separated but
  non-orthogonal, with inter-framework geometry reflecting known
  relationships. Mean pairwise cosine similarity in \((0, 1)\) with
  structured variation.
\end{enumerate}

\textbf{Effective dimensionality.} We compute the effective
dimensionality of the 6-direction set at each layer via PCA on
the \(6 \times 2048\) matrix of probe directions. Effective
dimensionality is the number of principal components explaining
\(\geq 90\%\) of variance. Low dimensionality (1--2) indicates
collapse; high dimensionality (5--6) indicates separation.

\textbf{Hierarchical clustering.} We apply Ward's method to the cosine
distance matrix (\(1 - \cos(\cdot, \cdot)\)) at each layer.
Dendrograms reveal whether the six foundations cluster into
interpretable groups.

\textbf{Permutation test for MFT group structure.} MFT predicts that
the six foundations divide into \emph{individualizing} (care, fairness,
liberty) and \emph{binding} (loyalty, authority, sanctity) clusters
\citep{graham2013mft}. We test this prediction with a permutation
test: compute the observed difference between mean within-group
cosine similarity and mean between-group cosine similarity, then
enumerate all \(\binom{6}{3} = 20\) assignments of the six foundations
to two groups of three as the exact null distribution, counting a
statistic within \(10^{-12}\) of the observed one as a tie. Each split
appears twice (either group can be listed first) with the same
statistic, so \(p\) is a multiple of \(0.10\) and the smallest attainable
value is \(p = 0.10\): the test cannot reject at \(\alpha = 0.05\).

\subsection{Bootstrap direction stability}\label{bootstrap-direction-stability}

With 32 training pairs per foundation in 2048 dimensions, the probe
has 2049 parameters and only 64 training examples. The classification
\emph{accuracy} may be robust in this regime (a single hyperplane suffices
for binary separation), but the extracted \emph{direction} could be noisy
, and noise in directions contaminates the angular analysis.

We assess direction stability via bootstrap resampling: for each
foundation at each layer, we resample the 32 training pairs with
replacement 200 times, retrain the probe on each bootstrap sample,
and compute the cosine similarity of each bootstrap direction with
the full-data direction. A mean bootstrap cosine similarity \(> 0.8\)
indicates a stable direction.

This gives a per-layer, per-foundation reliability metric for the
geometric analysis. Layers where directions are unstable should be
interpreted with caution.

\subsection{Framework-specific fragility}\label{framework-specific-fragility}

We extend the fragility protocol of
\citet{reblitzrichardson2026fragility} to per-foundation probes.
For each foundation at each layer:

\begin{enumerate}
\def\labelenumi{\arabic{enumi}.}
\tightlist
\item
  Train a linear probe (fixed seed) on clean activations.
\item
  Evaluate on clean test activations (baseline accuracy).
\item
  For each noise level \(\sigma \in \{0.1, 0.3, 1.0, 3.0, 10.0\}\)
  (extended to \(\{\ldots, 30, 100\}\) where censoring at \(\sigma=10\) is
  heavy, e.g.~the 7B run), add \(\mathcal{N}(0, \sigma^2)\) noise to
  cached test activations and re-evaluate, averaging accuracy over 10
  noise seeds.
\item
  The \textbf{critical noise} \(\sigma^*\) is the smallest \(\sigma\) where the
  seed-mean accuracy drops below 0.6. Following the convention of
  \citet{reblitzrichardson2026fragility}, a layer whose probe never
  drops below threshold is censored at the grid maximum (not dropped),
  and \(\sigma^*\) aggregates are means over all layers with a bootstrap
  CI over the noise seeds.
\end{enumerate}

We initially asked whether foundations differ in robustness (e.g.\\
whether care/harm is more robustly encoded than sanctity/degradation),
and whether the output dilution effect
\citep{reblitzrichardson2026dilution} is foundation-uniform. As \Cref{differential-fragility-across-frameworks}
reports, once accuracy is seed-averaged the per-foundation \(\sigma^*\)
values are not separable; the supported result is the uniform
cross-architecture difference, which an RMS-normalized control (\S4.11)
further attributes to activation scale.

\subsection{Probing dataset}\label{probing-dataset}

We use the same 240-pair minimal-pair probing dataset from
\citet{reblitzrichardson2026fragility}, now used at the
per-foundation level: 40 pairs per MFT foundation, stratified
80/20 into 32 train and 8 test pairs per foundation. Each pair
consists of a moral sentence tagged with its MFT foundation and a
matched neutral sentence controlling for sentence length, syntactic
structure, and topic. The dataset was drawn from a 1,200-pair
candidate pool generated by Claude Sonnet 4.6 from hand-written
seed examples, with automated validation gates (length ratio \(\leq
1.5\), embedding similarity, keyword scan, deduplication) and
LLM-as-judge filtering for neutral-pair quality.

For the geometric analysis, the 32 training pairs per foundation
yield \(64\) training examples (moral + neutral) for the
\texttt{nn.Linear(2048,\ 1)} probe. While this is a small sample for a
2049-parameter model, the bootstrap stability analysis (\Cref{bootstrap-direction-stability})
directly quantifies whether the resulting directions are reliable.

\subsection{Dilemma compositionality analysis}\label{dilemma-compositionality-analysis}

Experiments 1--7 establish that the model maintains distinct
directions for each moral foundation. A natural follow-up: when
two foundations \emph{conflict} in a moral dilemma, does the model
represent the dilemma as a composition of its component foundation
directions, or does it develop a qualitatively new representation?

\textbf{Dilemma dataset.} We generate 300 moral dilemma scenarios (20
per each of the \(\binom{6}{2} = 15\) foundation pairs) using
Claude Sonnet 4.6, with hand-written seed examples per pair. Each
scenario pits two specific foundations against each other (e.g.,
care vs.~authority: ``The nurse administered an unapproved
painkiller to a dying patient because following protocol meant
hours more agony''). Each dilemma text is paired with a matched
neutral sentence. All pairs pass automated validation gates
(length ratio, keyword scan, deduplication).

\textbf{Dilemma-specific probes.} For each of the 15 foundation pairs,
we train a binary linear probe (same architecture as \Cref{foundation-specific-probing}) to
distinguish dilemma moral text from matched neutral text. The
probe weight vector \(\hat{\mathbf{w}}_{\text{dilemma}}\) is the
direction in representation space that separates the dilemma
content from neutral content.

\textbf{Subspace membership score.} To measure compositionality, we
project each dilemma direction onto the 2D subspace spanned by its
two component foundation directions. Given component directions
\(\hat{\mathbf{w}}_A\) and \(\hat{\mathbf{w}}_B\) from Experiment 1,
we orthogonalize them via Gram--Schmidt and compute the fraction
of the dilemma direction's variance explained by this 2D subspace:

\[S = \|\text{proj}_{\text{span}(\hat{\mathbf{w}}_A, \hat{\mathbf{w}}_B)} \hat{\mathbf{w}}_{\text{dilemma}}\|^2\]

A membership score of \(S = 1\) indicates full compositionality
(the dilemma direction lies entirely within the component
subspace); \(S = 0\) indicates complete independence. The null
baseline for a random unit vector in \(\mathbb{R}^{2048}\) projected
onto a random 2D subspace has expected membership \(2/2048 \approx
0.001\). We estimate the empirical null distribution from 10,000
random unit vectors.

\textbf{Component balance.} We decompose the within-subspace projection
into components along \(\hat{\mathbf{w}}_A\) and the
Gram--Schmidt-orthogonalized \(\hat{\mathbf{w}}_B\). The balance
ratio (fraction of the projection along the first component)
measures whether the dilemma direction is dominated by one
foundation or draws equally on both. A ratio near 0.5 indicates
balanced composition.

\textbf{Shared-component geometry.} If dilemma representations are
partially compositional, dilemma pairs that share a foundation
component should have more similar probe directions than pairs
with no shared foundation. We test this by comparing the mean
cosine similarity between dilemma directions for pairs that share
a component (e.g., care--fairness and care--loyalty, which share
care) versus pairs with no overlap (e.g., care--fairness and
loyalty--sanctity).

\textbf{Cross-architecture consistency.} We repeat the dilemma probing
and subspace analysis on OLMoE-1B-7B to test whether the
compositionality structure is architecture-specific or general.

\section{Results}\label{results}

\subsection{Foundation-specific probe accuracy}\label{foundation-specific-probe-accuracy}

All six foundation-specific probes achieve perfect or near-perfect
accuracy across the full depth of OLMo-2 1B. Every foundation
reaches 100\% peak accuracy. Authority/subversion achieves 100\% at
all 16 layers; the remaining foundations show minor fluctuations at
individual layers (minimum 87.5\% for care/harm at layer 1). The
onset threshold (0.6) is exceeded at layer 0 for all foundations.

These results confirm that moral foundation content is linearly
separable from the earliest layer, consistent with the ``immediate
onset'' finding of \citet{reblitzrichardson2026fragility} for the
pooled binary moral/neutral probe. The per-foundation decomposition
shows no qualitative difference in \emph{detectability} across
foundations: all are equally easy to decode. The interesting
variation is not in accuracy but in the \emph{geometry} of the probe
directions that achieve this accuracy.

On OLMoE-1B-7B, all foundations similarly reach 100\% peak accuracy,
with the lowest early-layer values at 81.25\% (layers 0--1). The
accuracy profiles are essentially indistinguishable between
architectures.

These peak accuracies are maxima over 16 layers on small held-out sets
(8--16 examples per foundation), so they should be read as ``easily
decodable'' rather than as precise point estimates: the exact binomial
95\% confidence interval on a perfect 16/16 still reaches down to
\({\sim}0.79\). The geometric analysis below, not these saturated
accuracies, carries the paper's claims.

\subsection{Framework geometry: integration, not collapse}\label{framework-geometry-integration-not-collapse}

\begin{figure}[t]
\centering
\includegraphics[width=\linewidth]{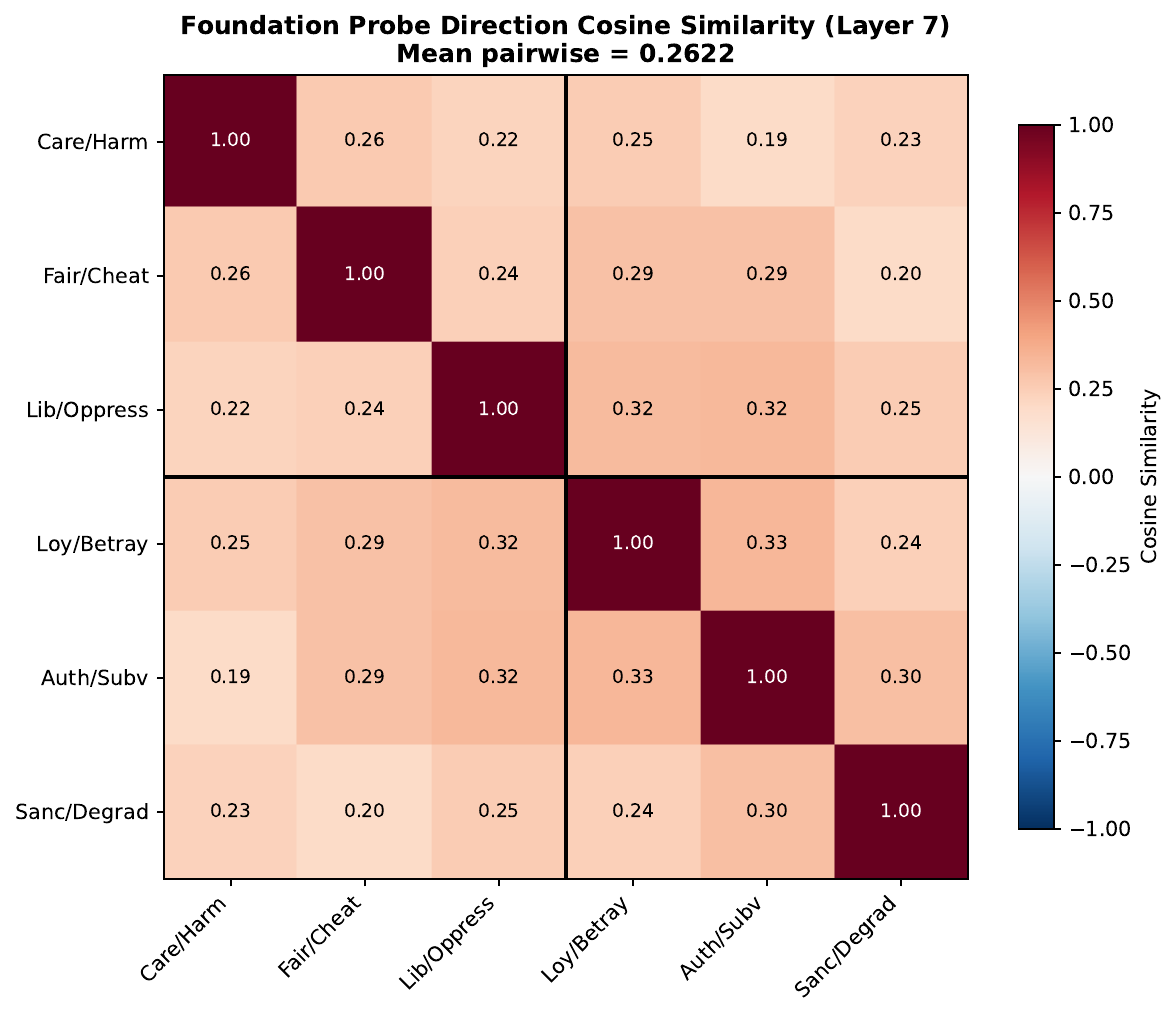}
\caption{Pairwise cosine similarity between foundation probe directions at layer 7 (bootstrap-stable; all six foundations exceed the 0.8 stability threshold at this layer; \Cref{direction-stability-under-bootstrap}), OLMo-2 1B. Mean off-diagonal cosine = 0.262. See Appendix \ref{app:cosine} for matrices at layers 0 and 15.}
\label{fig:cosine_heatmap}
\end{figure}

The headline finding: foundation probe directions are \emph{separated},
not collapsed. Across bootstrap-stable layers (6--15, where all six
directions exceed the 0.8 stability threshold; \Cref{direction-stability-under-bootstrap}, with the single
exception of authority at layer 8, 0.792), mean pairwise
cosine similarity ranges from \textbf{0.232 to 0.274}, far below the
collapse threshold (\(>0.95\)) and below the intermediate zone
(\(0.8\)--\(0.95\)). Figure \ref{fig:cosine_heatmap} shows the
representative pattern at layer 7 (mean cosine 0.262). The
peak-separation layer (layer 0, mean cosine 0.216) is below the
bootstrap stability threshold for all foundations and is reported
in Appendix \ref{app:cosine}.

The cosine similarities are uniformly \emph{positive} at all layers
(range 0.14--0.35), indicating that the foundation directions share
a common component. This is the \emph{integration} signature from our
trichotomy: the directions are separated but non-orthogonal,
consistent with a shared moral-salience subspace from which
foundation-specific directions deviate.

\textbf{Effective dimensionality.} The six foundation directions span
5 effective dimensions (the number of PCs explaining \(\geq 90\%\)
of variance) at every layer, confirming that the directions do not
collapse into a lower-dimensional subspace. The PCA is computed on the
mean-centered direction matrix, so for six directions its maximum is
5 (\(n-1\)); a value of 5 is therefore the ceiling, not a graded signal.
Effective dimensionality does not establish \emph{integration}: six random
directions in this hidden space also span \({\sim}5\) effective
dimensions (\Cref{geometric-trajectory-during-training}), so a near-maximal eff-dim is equally consistent with
the isolation regime. It rules out collapse, nothing more.

What separates integration from isolation is the uniformly positive
mean pairwise cosine (0.26, vs.~\({\sim}0\) for random directions). A
prior revision measured this only against a random-vector null and so
left open whether the shared component is moral-specific or a generic
content-vs-neutral axis that any six loaded-vs-neutral probes would
carry. We now calibrate it against a \emph{matched non-moral} battery: six
non-moral concept probes (sentiment, register, grammaticality, tense,
number, topic), built identically to the six foundations, same \(n = 32\)
training pairs, the same foundation-specific matched-twin neutrals, the
same probe-weight estimator, the same OLMo-2 1B and 16 layers. The
matched non-moral directions give a mean pairwise cosine of \textbf{0.013}
(bootstrap CI \([0.005, 0.020]\)), on the isotropic floor, while the six
moral foundations give \textbf{0.26}. The paired difference, resampling the
32 pairs, is \(\Delta = 0.223\), CI \([0.202, 0.244]\), excluding 0 (Table
\ref{tab:nonmoral_ladder}): the moral shared component is
\({\sim}20\times\) the matched non-moral null. This is a positive control
on the control: all six non-moral probes decode at 1.00 peak accuracy,
so their near-zero cosine is a genuine absence of a shared axis, not a
dead-probe artifact.

\begin{table}[t]
\centering
\caption{Calibration ladder for the mean pairwise cosine (OLMo-2 1B, layer 7, probe-weight directions, six concepts per construction; paired bootstrap over the 32 direction-estimation pairs, $n = 200$). The moral shared component sits ${\sim}20\times$ above the matched non-moral null and below the shared-neutral-pool construction. Mean-cosine entries are bootstrap means; the direct layer-7 moral point is 0.26 (Figure \ref{fig:cosine_heatmap}), the ${\sim}0.02$ gap to the bootstrap mean being resampling attenuation, which widens rather than narrows the moral$-$non-moral difference. PC1 columns give observed vs.\ the closed-form $[1+(k-1)\bar{c}]/k$.}
\label{tab:nonmoral_ladder}
\begin{tabular}{lccc}
\toprule
construction & mean cosine & PC1 (obs / pred) & peak acc \\
\midrule
isotropic floor & ${\sim}0$ & ${\approx}1/6$ (chance) & --- \\
matched non-moral battery & 0.013 [0.005, 0.020] & 0.183 / 0.179 & 1.00 \\
\textbf{six moral foundations} & 0.24 [0.22, 0.25] & 0.388 / 0.387 & 0.94--1.00 \\
shared-neutral-pool (non-moral) & 0.53 [0.51, 0.55] & 0.612 / 0.609 & 1.00 \\
\bottomrule
\end{tabular}
\end{table}

The reviewer's mechanism, a shared axis induced purely by contrasting
any loaded content against neutral text, is real, and we can exhibit it:
pooling six non-moral \emph{marked poles} against a single \emph{shared} set of 40
neutral statements, the construction that maximizes the generic
content-vs-neutral axis, gives a mean cosine of \textbf{0.53} (bootstrap CI
\([0.51, 0.55]\)), \emph{higher} than the moral 0.26. The moral and matched
non-moral probes avoid this inflation by construction: each pole is
contrasted against its own foundation-specific matched twin, which
cancels the shared neutral direction. That the moral number (0.26) sits
far below the shared-pool number (0.53) shows the estimator is not
manufacturing the moral shared component; the 0.26 is what survives
after the generic axis is differenced out.

The estimator dependence in \S4.13 (mean-difference 0.41 vs.\\
probe-weight 0.22) is consistent with this reading: the cosine magnitude
tracks how much of the moral-vs-neutral contrast each estimator retains,
and the 0.223 gap is reported on the probe-weight estimator used
throughout. The concentration of variance on a shared leading axis is
not independent corroboration. For \(k\) near-equicorrelated unit vectors
the first principal component captures \([1 + (k-1)\bar{c}]/k\) of the
variance; this identity is confirmed across all three constructions in
Table \ref{tab:nonmoral_ladder}, predicted and observed agreeing to
\(<0.01\) (moral 0.387/0.388, non-moral 0.179/0.183, shared-pool
0.609/0.612), which both anchors PC1 as a re-expression of the mean
cosine, not a second line of evidence, and validates the estimator.
Over the stable layers 6--15 the observed first-PC fraction is \textbf{0.379}
(\(\bar{c} = 0.26\)), vs.~\textbf{0.179} (\({\approx}1/6\)) for six random unit
vectors in 2048 dimensions.

The strongest remaining objection is that this battery spans affective,
syntactic, stylistic, and topical concepts but does not isolate whether
the shared moral axis is specifically \emph{moral} rather than generic
\emph{evaluative/affective} salience, since moral statements are also
emotionally charged. The discriminating control, a matched-twin
non-moral valence/affective battery, is named in \S5.6 and not run here.
Relative to a matched non-moral battery the shared component is
moral-specific; affective-vs-moral is the open residual.

\begin{figure}[t]
\centering
\includegraphics[width=\linewidth]{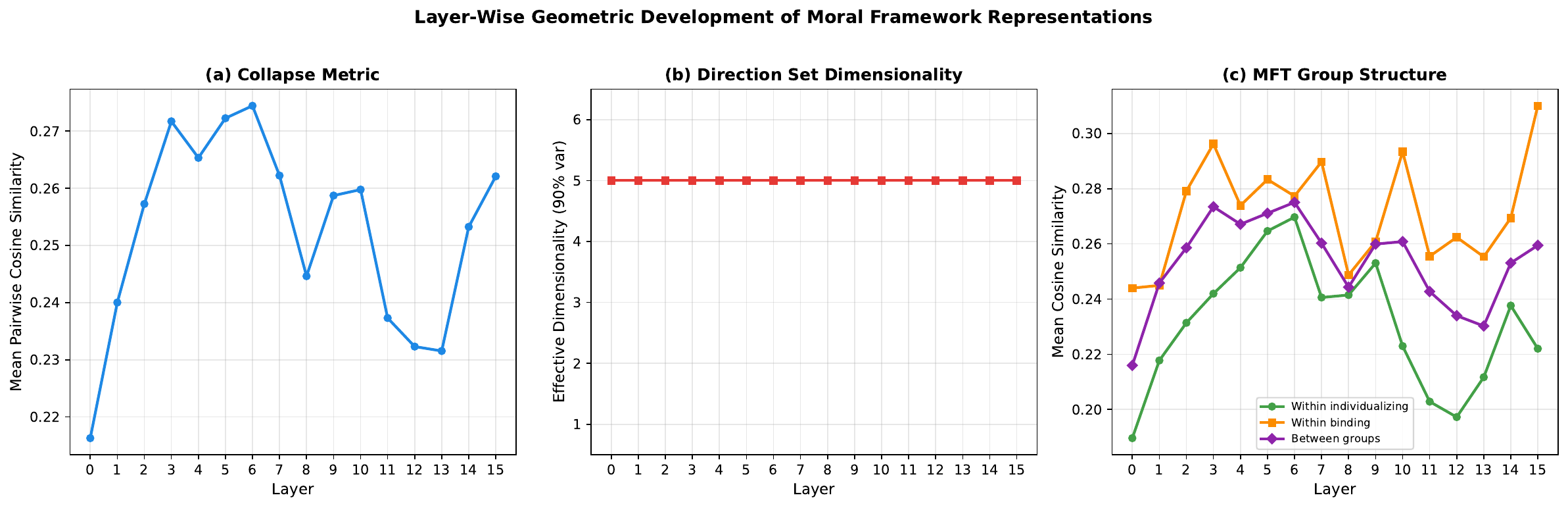}
\caption{Layer-wise geometric metrics for OLMo-2 1B. (a) Mean pairwise cosine similarity is relatively flat across layers. (b) Effective dimensionality remains constant at 5 across all layers. (c) MFT group structure: mean cosine within the individualizing group, within the binding group, and between groups track together across layers, so the directions do not separate into the predicted individualizing/binding clusters.}
\label{fig:layerwise}
\end{figure}

\subsection{Dendrogram structure does not recover MFT groups}\label{dendrogram-structure-does-not-recover-mft-groups}

\begin{figure}[t]
\centering
\includegraphics[width=0.7\linewidth]{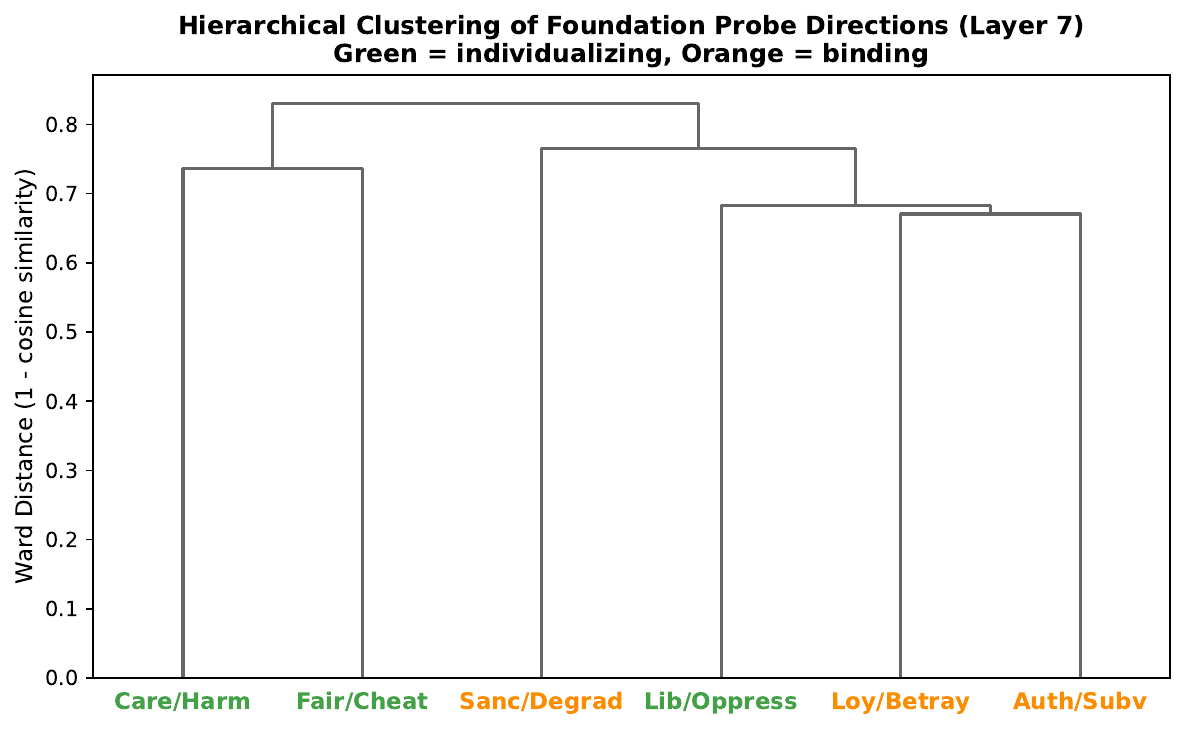}
\caption{Hierarchical clustering (Ward's method) of foundation probe directions at layer 7 (bootstrap-stable), OLMo-2 1B. The first split does not recover the MFT individualizing/binding distinction.}
\label{fig:dendrogram}
\end{figure}

Hierarchical clustering of the six foundation directions does
\emph{not} recover the MFT individualizing/binding distinction at any
layer. At layer 7, loyalty and authority merge first, liberty joins
them, then care and fairness merge separately, then sanctity joins
the loyalty--authority--liberty cluster. No layer produces the
predicted \{care, fairness, liberty\} vs.~\{loyalty, authority,
sanctity\} partition.

The most consistent clustering pattern across layers is a
care--sanctity pairing: these two foundations appear in the same
cluster at 10 of 16 layers. This crosses the MFT boundary (care is
individualizing, sanctity is binding) but has a plausible semantic
interpretation: both foundations concern protection of
vulnerable entities (persons from harm, sacred things from
degradation). The loyalty--authority pairing at layer 7 is within
the binding group, but liberty (individualizing) joins this cluster
before grouping with the other individualizing foundations (care,
fairness), further mixing MFT categories.

The permutation test for the individualizing/binding distinction
is non-significant at every layer (exact \(p = 0.40\) to \(0.80\), median
\(0.55\), against an attainable floor of \(0.10\); Appendix
\ref{app:permutation}). With 6 items there are only 10 distinct 3--3
splits, so the test cannot reject at \(\alpha = 0.05\) on any data; at
no layer does the MFT split rank above fourth of the ten. Together
with the absence of MFT-aligned dendrograms at any layer, this
indicates that the model's inter-framework geometry does not reflect
the MFT group structure on this dataset.

\subsection{Layer-wise geometric development}\label{layer-wise-geometric-development}

The geometric structure is relatively stable across layers. Mean
pairwise cosine similarity is lowest at layer 0 (0.216), rises
modestly to a peak of 0.274 at layer 6, then returns to 0.232 at
layer 13 before a slight increase at layer 15 (0.262). The
variation is small (range 0.06) compared to the distance from
collapse, indicating that framework separation is a consistent
property of the representation space rather than an emergent
late-layer phenomenon.

Effective dimensionality remains constant at 5 across all layers,
meaning the rank of the direction set does not change even as the
pairwise angles shift. The geometric structure is a rotation of
the direction set within a fixed-rank subspace, not a
dimensionality change.

\subsection{Dense vs.~MoE framework geometry}\label{dense-vs.-moe-framework-geometry}

Framework geometry is remarkably similar between architectures.
OLMoE-1B-7B shows mean pairwise cosine similarity ranging from
0.219 to 0.287 across layers, compared to OLMo-2 1B's 0.216 to 0.274
(the dense side reuses the \Cref{layer-wise-geometric-development} exp1 directions, so this range is the
same one reported there; the two architectures' directions are
estimated by the identical probe-weight procedure). Effective
dimensionality is 5 at all layers for both models. The overall
degree of framework separation is consistent across dense and MoE
architectures.

Neither architecture produces MFT-aligned dendrogram structure at
any layer. Both models show the care--sanctity pairing as the
most stable clustering feature, suggesting that inter-framework
geometry is driven by semantic relationships in the training corpus
rather than by architectural properties. The permutation test for
MFT group structure is non-significant at all layers for both models
(exact \(p \geq 0.40\) dense, \(\geq 0.30\) MoE; floor \(0.10\)).

This finding extends \citet{reblitzrichardson2026dilution}: output
dilution affects moral encoding \emph{scale} (74\(\times\) signal gap) but
not framework \emph{structure}. The geometric organization of moral
foundations is preserved across architectures.

\subsection{Direction stability under bootstrap}\label{direction-stability-under-bootstrap}

Bootstrap resampling (200 iterations) shows a stability gradient
across layers. Early layers (0--5) show borderline stability (mean
cosine similarity with the full-data direction: 0.74--0.80, below
the 0.8 threshold for most foundations). Middle and late layers
(6--15) are stable (mean cosine \(> 0.80\)). Sanctity/degradation
is the most stable foundation (13/16 layers stable); care/harm the
least (10/16 stable).

This gradient has two implications. First, the geometric analysis
at early layers, including the peak separation layer (layer 0),
should be interpreted with caution, as the specific pairwise
cosine values may shift under resampling. Second, the stability
gradient itself is informative: probe directions become more
determined as representations become more specialized, paralleling
the lexical-to-compositional gradient from
\citet{reblitzrichardson2026fragility}.

The headline geometric findings (separation not collapse; effective
dimensionality = 5) are confirmed at layers 6--15 where directions
are stable.

\subsection{Differential fragility across frameworks}\label{differential-fragility-across-frameworks}

\begin{figure}[t]
\centering
\includegraphics[width=\linewidth]{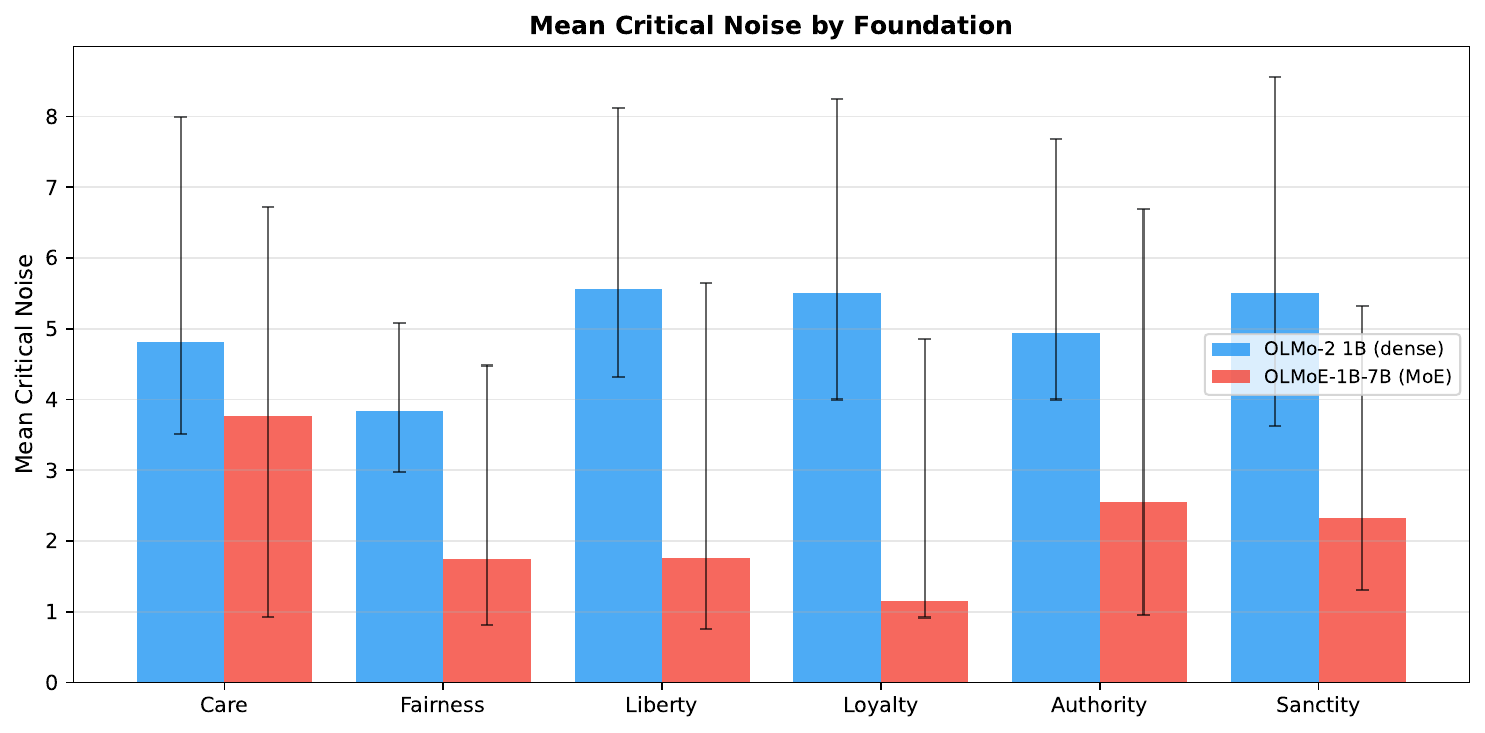}
\caption{Mean critical noise $\sigma^*$ per foundation for OLMo-2 1B (left) and OLMoE-1B-7B (right), seed-averaged with cap-at-max and bootstrap 95\% CIs over noise seeds. Every foundation is more fragile in MoE than in dense (the cross-architecture dilution effect); within each architecture the per-foundation CIs overlap and the ordering is not statistically separable.}
\label{fig:fragility}
\end{figure}

We measure per-foundation critical noise \(\sigma^*\) with the standard
fragility protocol, averaging accuracy over 10 noise seeds and
censoring never-fragile layers at the grid maximum (the cap-at-max
convention of \citet{reblitzrichardson2026fragility}); bootstrap 95\%
CIs are over the noise seeds. Seed-averaging is necessary here: a
single noise draw makes per-layer \(\sigma^*\) unstable.

\textbf{The result is between architectures, not between foundations.}
Every foundation is more fragile in OLMoE than in OLMo-2: dense
\(\sigma^*\) runs 3.8--5.6 across the six foundations (mean 5.0), MoE
\(\sigma^*\) runs 1.2--3.8 (mean 2.2), a per-foundation architecture gap
of \({\sim}2.3\times\). This is the same direction as, and smaller than,
the 4.2\(\times\) pooled gap reported by
\citet{reblitzrichardson2026dilution}; the pooled probe trains on
192 pairs versus 32 per foundation here, so it has more statistical
power and a cleaner separation. Output dilution suppresses moral
encoding across the board. This cross-architecture gap is itself the
output-scale effect: an RMS-normalized control (scaling noise to each
layer's activation RMS; \citealp{reblitzrichardson2026fragility}, \Cref{layer-wise-geometric-development})
shrinks it from \({\sim}2.3\times\) to \({\sim}1.2\times\) (dense mean
\(\sigma^* = 10.0\), MoE \(8.2\) at the matched grid), exactly as expected
if the MoE's smaller output scale, not weaker encoding, drives the raw
fragility difference.

\textbf{Within an architecture, foundations are not reliably separable.}
The per-foundation \(\sigma^*\) values have wide, overlapping bootstrap
CIs (dense: e.g.~sanctity \(5.50\), CI \([3.6, 8.6]\); care \(4.81\), CI
\([3.5, 8.0]\). MoE: sanctity \(2.33\), CI \([1.3, 5.3]\); care \(3.76\), CI
\([0.9, 6.7]\)), and no foundation is robustly most or least fragile.
The binding-vs-individualizing group difference is not significant in
either model and reverses sign between them (dense: binding \(-\)
individualizing \(= +0.58\), exact permutation \(p = 0.40\); MoE: \(-0.41\),
\(p = 0.70\)). We therefore make no per-foundation or MFT-group fragility
claim; the supported finding is the uniform cross-architecture dilution.

\subsection{Geometric trajectory during training}\label{geometric-trajectory-during-training}

\begin{figure}[t]
\centering
\includegraphics[width=\linewidth]{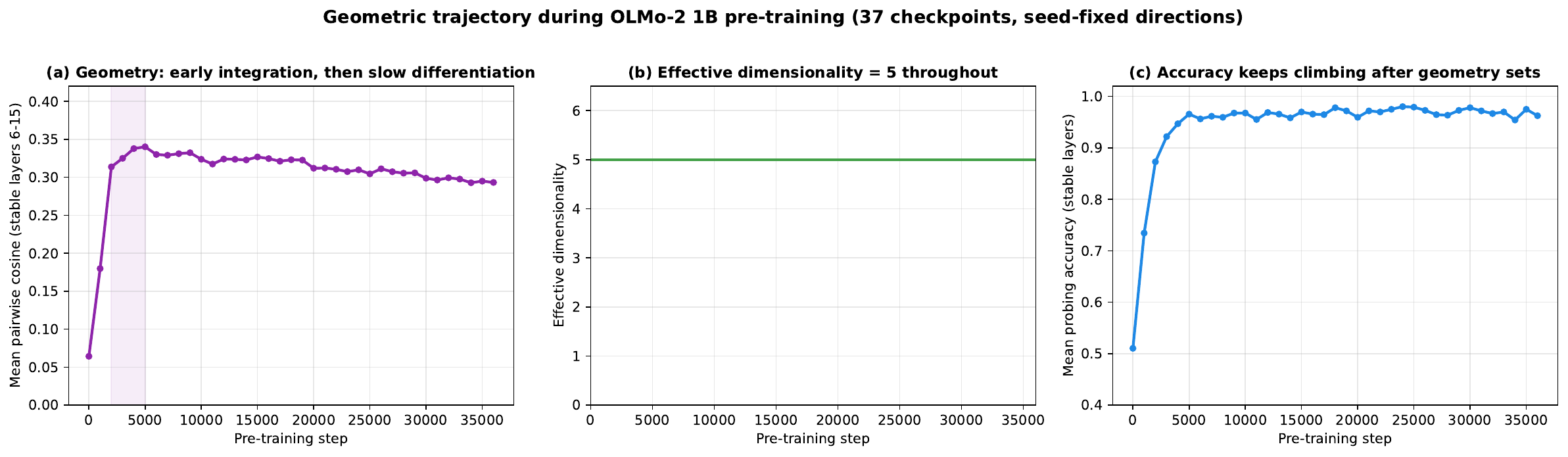}
\caption{Geometric trajectory during OLMo-2 1B pre-training (37 checkpoints, seed-fixed foundation directions). (a) Mean pairwise cosine reaches the integration regime by step 2--5K, then slowly differentiates. (b) Effective dimensionality remains at 5 throughout. (c) Accuracy keeps climbing after the geometry sets.}
\label{fig:trajectory}
\end{figure}

Framework geometry reaches its integration regime early and then
slowly differentiates. We track the mean pairwise cosine similarity of
the six foundation directions (averaged over the stable layers 6--15)
across all 37 OLMo-2 1B early-training checkpoints (step 0 to 36,000).
The directions are estimated with a fixed probe seed; an earlier
unseeded run that was resumed in batches produced an RNG-dependent
artifact (a spurious 0.38/0.28 bimodality with spikes at 5K-multiple
checkpoints), which the seeded recomputation removes while leaving the
smooth accuracy trajectory unchanged.

\begin{itemize}
\item
  \textbf{Step 0.} Mean cosine is 0.06; the six foundation directions are
  nearly orthogonal, as expected for probes on random representations.
  Mean accuracy is 0.51 (chance).
\item
  \textbf{Step 1000.} Cosine jumps to 0.18; accuracy reaches 0.73. Shared
  moral-salience structure has begun to form.
\item
  \textbf{Steps 2000--5000.} Cosine rises to its peak (0.31 at step 2K,
  0.34 at step 5K) as accuracy climbs from 0.87 to 0.97. The
  integration regime is reached here, while accuracy still has \({\sim}10\)
  points to gain.
\item
  \textbf{Steps 5000--36,000.} Cosine then slowly \emph{declines}, from 0.34 to
  0.29 (\({\sim}13\%\)), as accuracy holds at 0.96--0.98. The foundation
  directions, having entered the integration band early, keep
  differentiating gradually for the rest of the observed training.
\end{itemize}

Effective dimensionality remains constant at 5 from step 0 onward.
This is expected: random unit vectors in \(\mathbb{R}^{2048}\) are
nearly orthogonal with high probability, so even at initialization
the six probe directions span 5 effective dimensions. The informative
signal is cosine similarity, which transitions from \(\approx 0\)
(random) into the integration band (\({\sim}0.3\)) within the first few
thousand steps, then drifts down slowly. The final-trajectory value
(0.29) is consistent with the fully-trained model (\Cref{framework-geometry-integration-not-collapse}, mean
\({\approx}0.26\) over stable layers): geometry sets its structure early
and refines it gradually rather than abruptly.

The temporal dissociation (framework geometry stabilizing at step
2000 while accuracy continues improving through step 25,000)
extends the two-phase pattern identified by
\citet{reblitzrichardson2026fragility}: accuracy saturates early, but
fragility continues resolving. Here we add a third metric:
inter-framework \emph{structure} also stabilizes before inter-framework
\emph{discriminability} finishes developing. The model discovers the
geometric layout of moral concepts early and then spends the
remainder of training strengthening the representations within that
fixed layout.

\subsection{Dilemma compositionality: partial but structured}\label{dilemma-compositionality-partial-but-structured}

\begin{figure}[t]
\centering
\includegraphics[width=\linewidth]{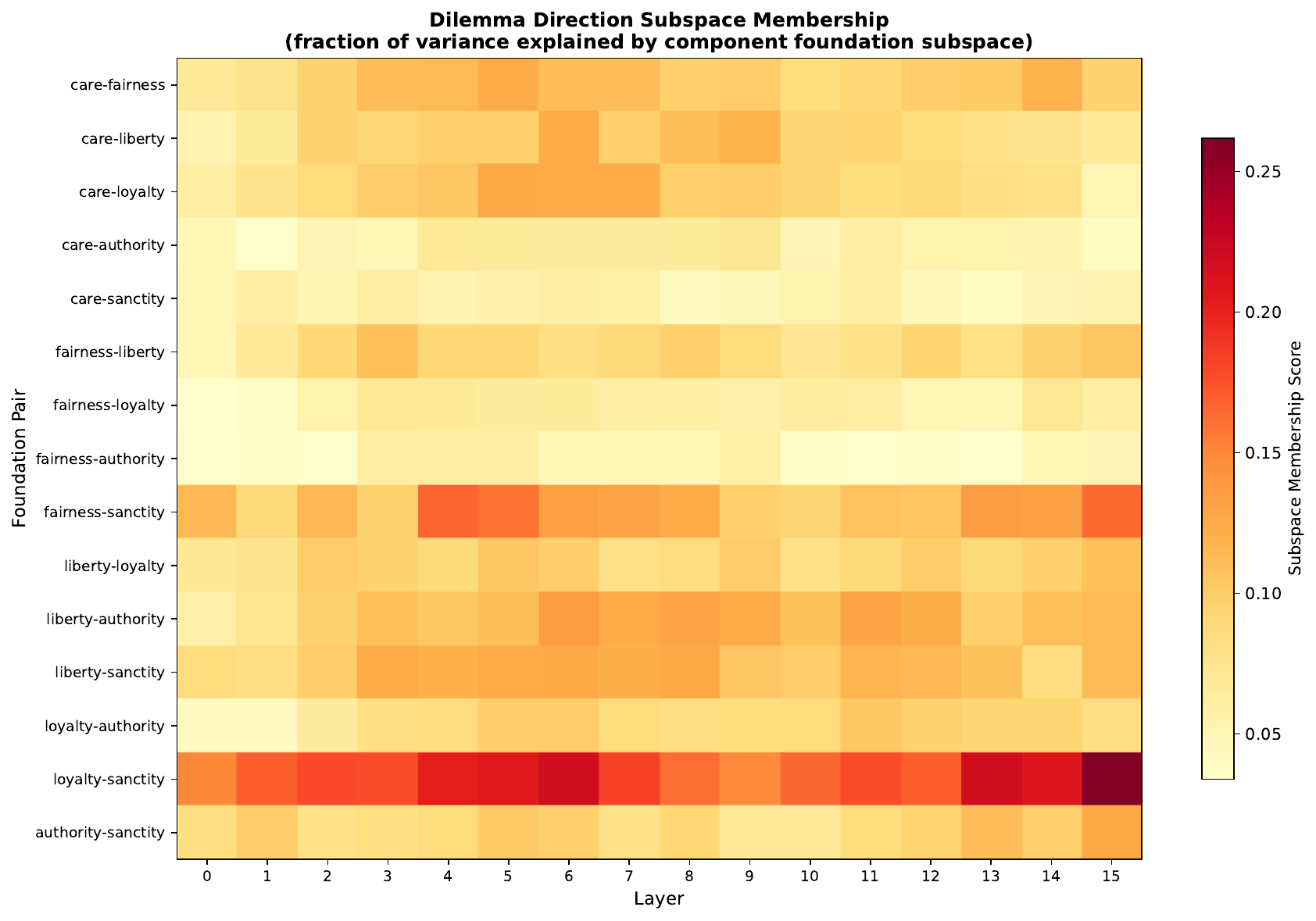}
\caption{Subspace membership scores for 15 dilemma probe directions across 16 layers. Each cell shows the fraction of a dilemma direction's variance explained by the 2D subspace of its component foundation directions. Liberty--sanctity shows the strongest compositional signal.}
\label{fig:dilemma_heatmap}
\end{figure}

We now ask whether the model's representation of moral \emph{dilemmas}
(scenarios where two foundations conflict) can be decomposed in
terms of the single-foundation directions from Experiment 1.

\textbf{Dilemma probes achieve high accuracy.} All 15 dilemma-specific
probes achieve \(\geq 75\%\) peak test accuracy (mean 94.2\%), with
13 of 15 pairs at \(\geq 87.5\%\). As with the foundation accuracies
(\Cref{foundation-specific-probe-accuracy}), these are maxima over 16 layers on small held-out sets
(20 pairs per dilemma, 80/20 split, so \({\sim}4\) test examples each), so
they should be read as ``easily decodable'' rather than as precise point
estimates. The model reliably distinguishes dilemma moral content from
matched neutral text.

\textbf{Subspace membership: partial compositionality.} The mean peak
subspace membership score across all 15 pairs is \textbf{0.118}. The right
null is not a random vector (mean 0.001), which ignores the shared
moral-salience component every moral direction carries, but the
\emph{mismatched-pair baseline}: the membership of each dilemma direction
in the 2D spans of foundation pairs it shares no component with. That
baseline is \textbf{0.044} at the matched peak layer, so the compositional
signal is \({\sim}2.7\times\) the mismatched baseline (and matched
exceeds mismatched at every layer; Appendix \ref{app:dilemma_membership}).
The matched\(-\)mismatched gap is \(0.074\), CI \([0.053, 0.100]\), excluding 0
(paired bootstrap over the 15 dilemmas, \(n = 10^4\)). This per-pair-peak
figure is a max-over-layers extremum and is biased upward; the unbiased
cross-layer-mean gap is \(0.052\), CI \([0.037, 0.069]\), also excluding 0
(matched 0.091 vs.~mismatched 0.039).
The signal is real but modest: even matched, only \({\sim}12\)\% of each
dilemma direction's variance is explained by its component foundation
subspace. The remaining \({\sim}88\)\% (mean residual norm = 0.939) lies
in directions orthogonal to both component foundations.

\begin{figure}[t]
\centering
\includegraphics[width=\linewidth]{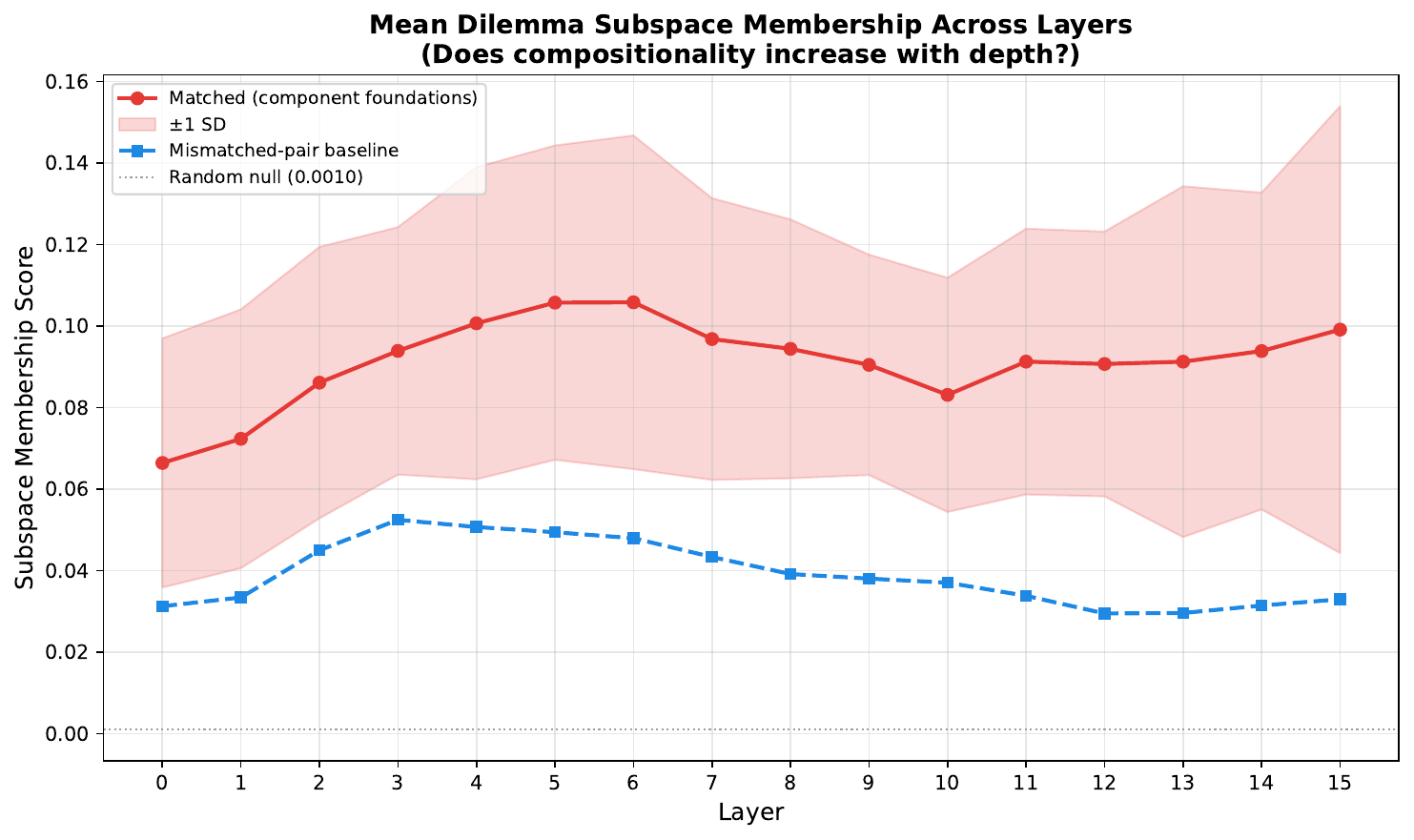}
\caption{Mean dilemma subspace membership across layers: matched (component foundations, red, $\pm 1$ SD band), the mismatched-pair baseline (blue dashed), and the random-vector null (gray, ${\sim}0.001$). Matched membership exceeds the mismatched baseline at every layer; both sit far above the random null.}
\label{fig:dilemma_layers}
\end{figure}

\textbf{Component balance is near-equal.} The mean component balance
ratio is 0.486 (perfect balance = 0.5). In 14 of 15 pairs, the
balance falls within {[}0.40, 0.58{]}, indicating that both component
foundations contribute approximately equally to the within-subspace
projection. The exception is fairness--sanctity (balance = 0.335),
where the sanctity component dominates. When
two foundations \emph{do} compose, they compose symmetrically rather
than one foundation dominating the representation.

\begin{figure}[t]
\centering
\includegraphics[width=\linewidth]{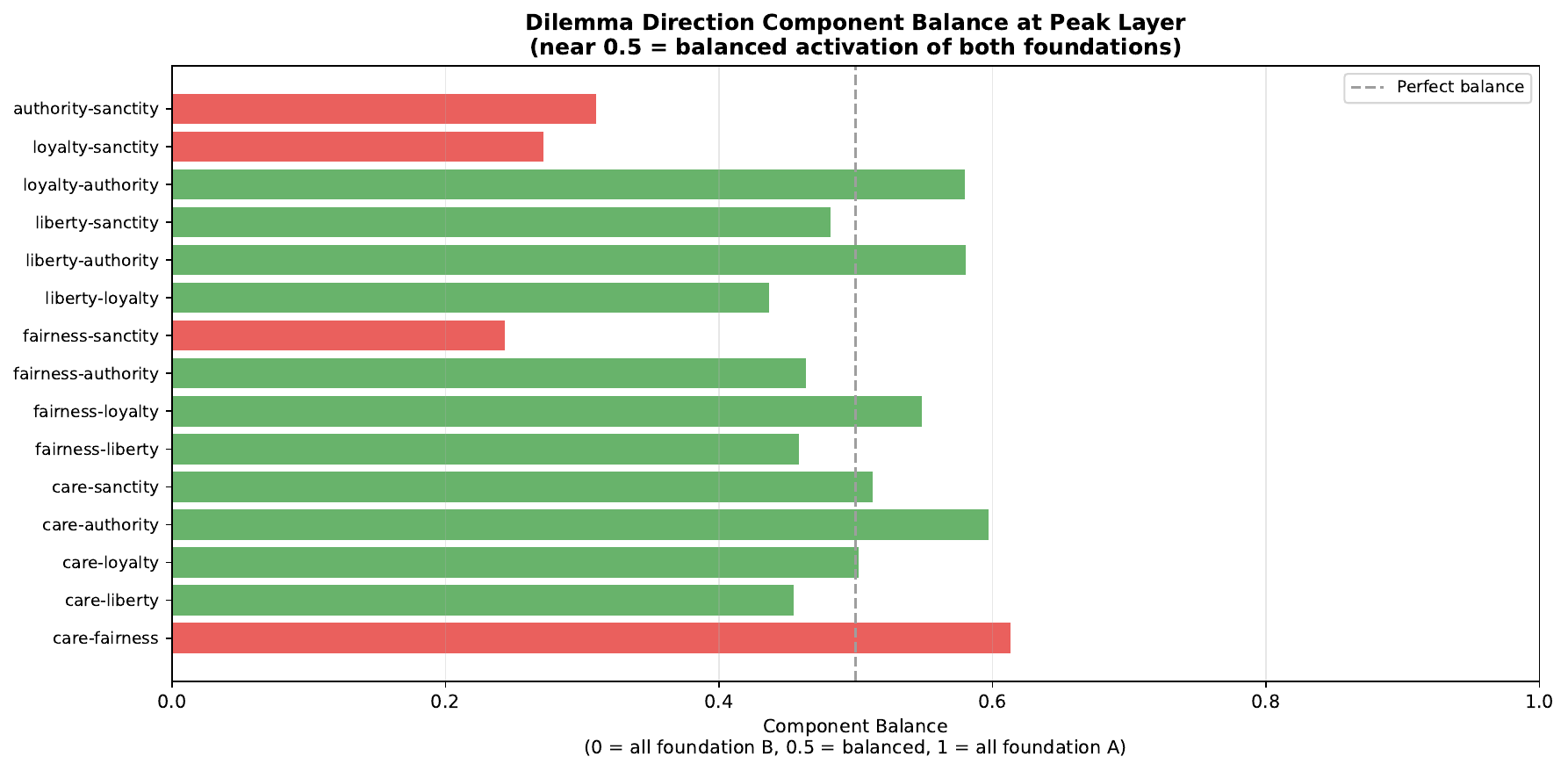}
\caption{Component balance at each pair's peak subspace membership layer. Values near 0.5 (dashed line) indicate balanced contribution from both foundations. Fairness--sanctity (red) is the only pair with substantial imbalance.}
\label{fig:dilemma_balance}
\end{figure}

\textbf{Shared-component structure.} Dilemma pairs that share a
foundation component have consistently higher cosine similarity
than pairs with no shared foundation. At the peak effect layer
(layer 13), the mean cosine similarity between shared-component
pairs is \textbf{0.273} versus \textbf{0.196} for non-sharing pairs (difference
= 0.076, CI \([0.035, 0.114]\), excluding 0; paired bootstrap over the 15
dilemmas, \(0.04\)\% of resamples \(\leq 0\)). A permutation test confirms
the effect: permuting which
foundation-pair label is attached to each dilemma direction (which
shuffles the share/no-share partition of the 105 pairwise cosines)
gives \(p = 0.0001\) at layer 13, and the minimum \(p\) across all layers
is also \(0.0001\) (10,000 permutations). The difference is positive at
every layer, so the compositional structure is a general property of
the dilemma direction geometry, not a layer-specific artifact.

\begin{figure}[t]
\centering
\includegraphics[width=0.9\linewidth]{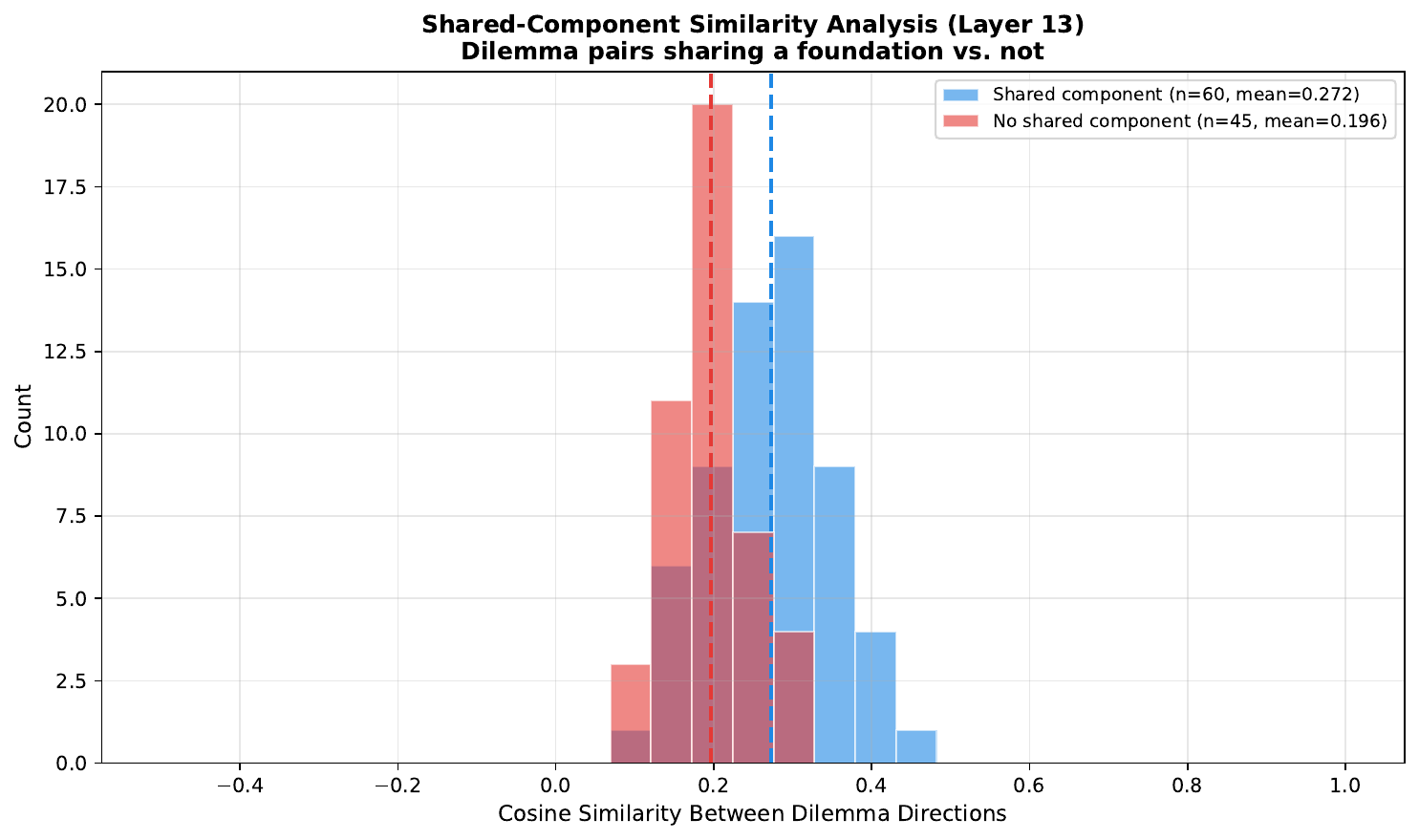}
\caption{Distribution of pairwise cosine similarities between dilemma directions at layer 13, split by whether the pairs share a component foundation. Shared-component pairs (blue, $n = 60$) have higher mean similarity (0.273) than non-sharing pairs (red, $n = 45$, mean 0.196); exact permutation $p = 0.0001$.}
\label{fig:shared_component}
\end{figure}

\textbf{Hierarchical clustering.} The 21-direction dendrogram (6
foundation + 15 dilemma directions) at layer 13 shows two
features. First, the six foundation directions form a distinct
cluster separate from the dilemma directions. Second, within the
dilemma cluster, pairs sharing a component foundation tend to merge
at lower distances, consistent with the shared-component analysis
above.

The categorical foundation/dilemma separation initially appears to
undercut the compositionality narrative: if dilemmas were purely
compositional, they would cluster near their component foundations,
not in a separate region. However, this separation is driven by
register features, not moral content. Projecting all 21 directions
into the 5D moral subspace (spanned by the six foundation
directions) and re-clustering dissolves the separation entirely:
foundations now cluster with their related dilemmas (e.g., sanctity
with fairness--sanctity and care--sanctity). The first-order
separation in the full-space dendrogram reflects the \({\sim}90\)\%
extra-moral residual (likely text-register differences between
declarative single-foundation sentences and narrative dilemma
scenarios), while the second-order structure within each cluster
reflects genuine moral content relationships.

\begin{figure}[t]
\centering
\includegraphics[width=\linewidth]{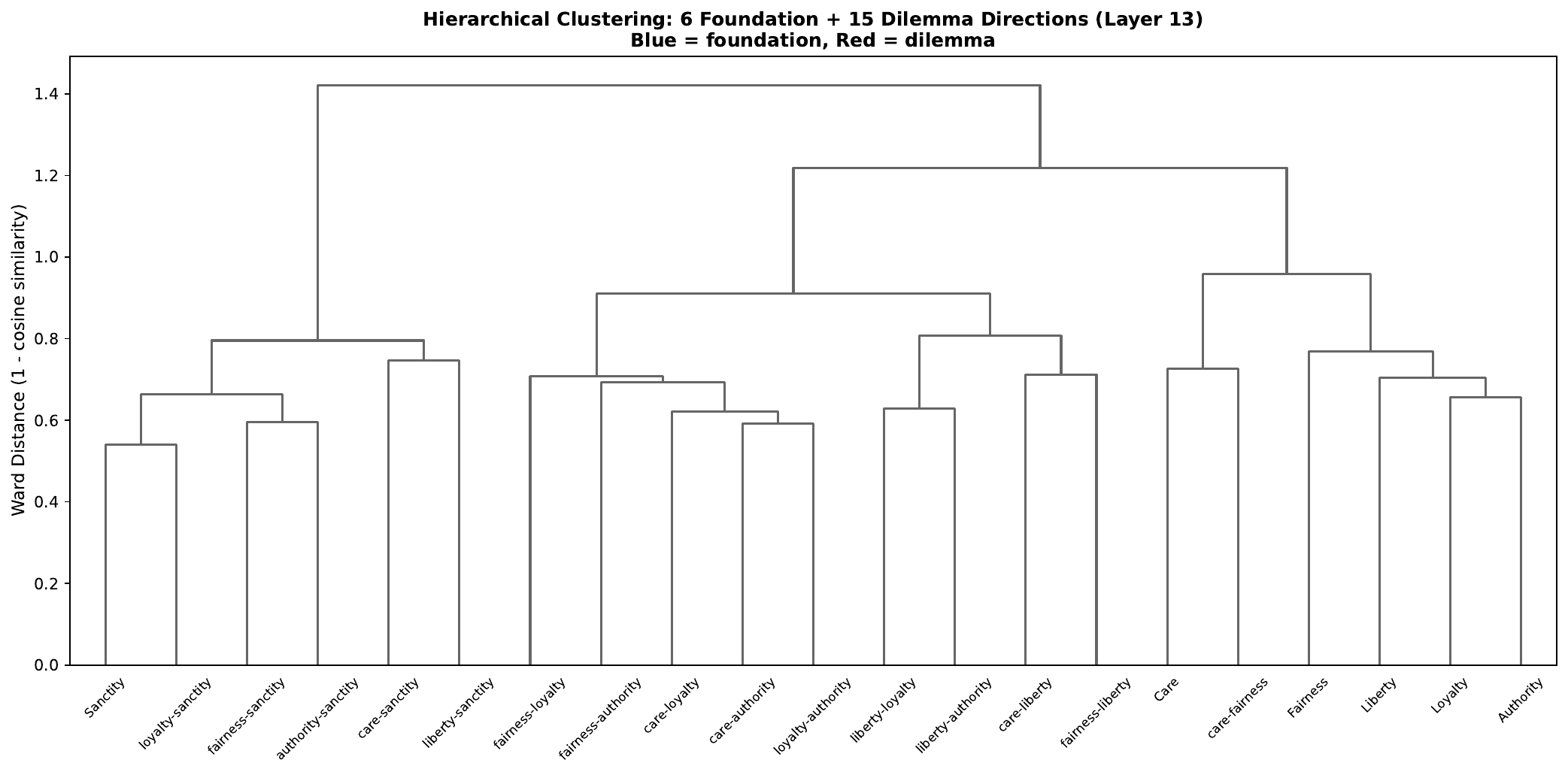}
\caption{Hierarchical clustering of all 21 directions (6 foundation in blue, 15 dilemma in red) at layer 13. Foundation directions cluster separately from dilemma directions.}
\label{fig:dilemma_dendrogram}
\end{figure}

\textbf{Full moral subspace projection.} The 2D subspace analysis above
uses only the two component foundations per dilemma. To test whether
dilemma representations are compositional over the \emph{full} moral
vocabulary, we project each dilemma direction onto the 5D subspace
spanned by all six foundation directions (5D because the six
directions have effective dimensionality 5). Averaging across all
layers and pairs, the mean full-subspace membership is \textbf{0.109}, only
modestly above the cross-layer 2D matched membership of 0.091 (ratio
1.19\(\times\); both far above the random null of \({\sim}0.001\), and the
2D matched value is \({\sim}2.3\times\) its mismatched-pair baseline of
0.039). The small gain from the 2D component subspace to the full moral
subspace indicates that the \({\sim}90\)\% residual is not explained by
\emph{any} foundation direction: it is genuinely extra-moral, likely
encoding conflict-specific features such as trade-off framing and
tension that lie outside the moral subspace entirely.

\textbf{Scale replication.} The partial-compositionality result reproduces
on OLMo-2 7B with the same matched-vs-mismatched structure. Mean peak
matched membership is 0.090, against a mismatched-pair baseline of
0.032 (\({\sim}2.8\times\), vs.~the \(2.7\times\) at 1B); mean residual
norm is 0.953; and component balance is 0.455 (vs.~0.486 at 1B, where
0.5 is perfect balance). The 7B model decomposes
moral dilemmas the same way the 1B model does: a small, significant
projection onto the component foundations, a large conflict-specific
residual, and near-balanced loading on the two foundations in tension.

\subsection{Dilemma direction stability}\label{dilemma-direction-stability}

Bootstrap resampling (50 iterations) of the 15 dilemma probe
directions across 16 layers yields 240 direction--layer
combinations. Of these, \textbf{239 are stable} (mean cosine with
full-data direction \(> 0.7\)), with the single exception at
liberty--loyalty, layer 0. The relaxed threshold (0.7 vs.~0.8 for
foundation probes) reflects the smaller sample size (16 training
pairs vs.~32 for foundations), but the results indicate that the
dilemma probe directions, and therefore the subspace analysis
built on them, are reliable.

\subsection{Complexity--fragility gradient}\label{complexityfragility-gradient}

\begin{figure}[t]
\centering
\includegraphics[width=0.7\linewidth]{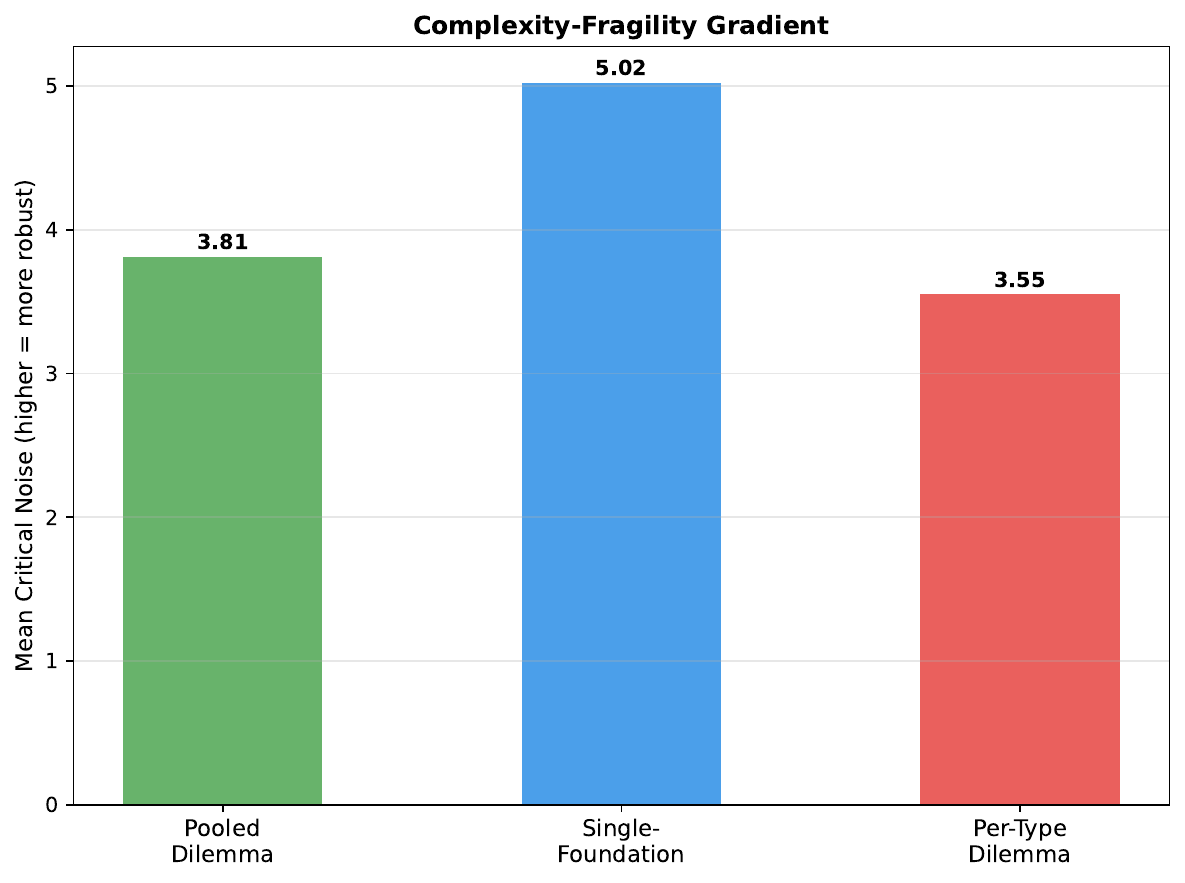}
\caption{Raw mean critical noise for probes at three complexity levels (single-foundation from Exp. 7, pooled and per-type dilemma). The apparent gradient does not survive RMS normalization (see text): under scale correction the single-foundation and pooled-dilemma values converge, indicating the raw ordering reflects register-dependent activation scale rather than encoding robustness.}
\label{fig:fragility_gradient}
\end{figure}

We compare fragility across three levels of moral complexity:
single-foundation probes (from Experiment 7), pooled binary
dilemma probes (all 300 dilemma pairs pooled), and per-type
dilemma probes (15 separate probes, 20 pairs each).

In raw activation units the mean critical noise looks like a
\textbf{complexity--fragility gradient}: single-foundation probes most
robust (\(\sigma^* = 5.02\)), pooled dilemma less (\(3.81\)), per-type
dilemma least (\(3.55\)), seemingly because more specific distinctions
are encoded with less redundancy. This gradient does not survive scale
normalization. Consistent with the activation-scale confound identified
in the companion work (\citealp{reblitzrichardson2026fragility}, \Cref{layer-wise-geometric-development}),
we re-evaluate fragility with noise scaled to each layer's activation
RMS. The single-foundation and pooled-dilemma \(\sigma^*\) then converge
(both at the grid maximum) and the per-type value barely separates:

{\def\LTcaptype{none} 
\begin{longtable}[]{@{}lrr@{}}
\toprule\noalign{}
Probe (1B dense) & Raw \(\sigma^*\) & RMS-normalized \(\sigma^*\) \\
\midrule\noalign{}
\endhead
\bottomrule\noalign{}
\endlastfoot
Single-foundation & 5.02 & 10.0 \\
Pooled dilemma & 3.81 & 10.0 \\
Per-type dilemma & 3.55 & 9.31 \\
\end{longtable}
}

The raw gradient therefore reflects register-dependent activation norms
(the narrative dilemma stimuli carry smaller-scale activations than the
declarative foundation pairs; cf.~\S5.5) rather than differential
encoding robustness. We retain raw \(\sigma^*\) only as a measure of
practical perturbation sensitivity and make no complexity-robustness
claim. The compositionality evidence in \S4.9--4.10 (subspace membership,
shared-component geometry, balanced loading) is geometric, not
fragility-based, and is unaffected.

\subsection{MoE architecture preserves compositionality}\label{moe-architecture-preserves-compositionality}

The dilemma probing and subspace analysis on OLMoE-1B-7B produces
results consistent with the dense model and the same matched-vs-mismatched
structure. Mean peak accuracy is 95.8\% (vs.~94.2\% on OLMo-2), mean
peak matched membership is \textbf{0.118} against a mismatched-pair baseline
of \textbf{0.045} (\({\sim}2.6\times\), matching the \(2.7\times\) on OLMo-2 1B).
The partial compositionality structure (a consistent matched-over-mismatched
margin at low absolute membership, with a large residual) is a property of
the representation geometry, not a dense-architecture artifact.

\subsection{Robustness to direction-finding method}\label{robustness-to-direction-finding-method}

The geometric findings reported above rely on probe weight vectors
as foundation directions. To test whether the geometry is an
artifact of discriminative probe training, we extract directions
using two alternative methods and compare.

\textbf{Mean-difference directions.} For each foundation, we compute
\(\mathbf{d}_f = \overline{\mathbf{a}}_{\text{moral}} -
\overline{\mathbf{a}}_{\text{neutral}}\) (the normalized difference
of class-conditional activation means), requiring no optimization.
These training-free directions replicate the core geometric
findings: effective dimensionality is 5 at all layers. Like the
probe-weight directions, the mean-diff dendrogram does not recover
the MFT individualizing/binding split at any layer, and the
permutation test is non-significant throughout. Per-foundation
cosine similarity between probe-weight and mean-difference
directions ranges from 0.67 to 0.72 (mean across layers),
indicating related but non-identical directions: the probe-weight
method finds more foundation-specific discriminative signal, while
the mean-difference method captures more of the shared
moral-salience component (mean pairwise cosine 0.41 vs.~0.22 at
layer 0).

\textbf{Representation-engineering directions.} We also test
paired-difference PCA \citep{zou2023representation}: for each
pair \(i\), compute \(\mathbf{d}_i = \mathbf{a}_{\text{moral},i} -
\mathbf{a}_{\text{neutral},i}\) and take the first principal
component. This method performs poorly: the first PC explains only
8--11\% of variance (barely above chance in \(\mathbb{R}^{2048}\)),
and the resulting directions show low alignment with probe-weight
directions (\(|\cos| = 0.07\)--\(0.26\)) and weak classification
accuracy. With \({\sim}32\) pairs per foundation in 2048 dimensions,
the \(p \gg n\) regime prevents PCA from isolating the concept
direction. This negative result validates that the convergent
probe-weight and mean-difference findings are not trivially
recoverable; they depend on direction-finding methods with
appropriate inductive bias for small datasets.

\textbf{Cross-register transfer.} Both probe-weight and mean-difference
directions transfer to narrative dilemma text with \(>90\)\% mean
pair accuracy across foundations (the fraction of pairs where the
direction projects the moral text higher than the neutral text).
Probe-weight directions achieve \(>95\)\% for all foundations; the
mean-difference method drops to \({\sim}91\)\% for authority/subversion,
the foundation with weakest directional stability. The transfer
gap between same-register (declarative test pairs) and cross-register
(dilemma pairs) averages under 4 percentage points, with
authority/subversion showing the largest gap (\({\sim}9\) pp for
mean-difference; \S5.5).

\subsection{Scale validation: geometry persists from 1B to 7B}\label{scale-validation-geometry-persists-from-1b-to-7b}

We replicate the core geometric findings on OLMo-2 7B (the 1124
release, 32 layers, hidden dimension 4096), a dense model with 7.3B
parameters, versus 1.5B for OLMo-2 1B. The integration signature holds.
Mean pairwise cosine similarity ranges from \textbf{0.193 to 0.244} across
layers (1B: 0.216--0.274), positive everywhere and far below the
collapse threshold. Effective dimensionality is \textbf{5 at every one of
the 32 layers}, identical to the 1B and MoE models. Foundation
directions are geometrically distinct, sharing a common moral-salience
component, at both scales.

\begin{figure}[t]
\centering
\includegraphics[width=\linewidth]{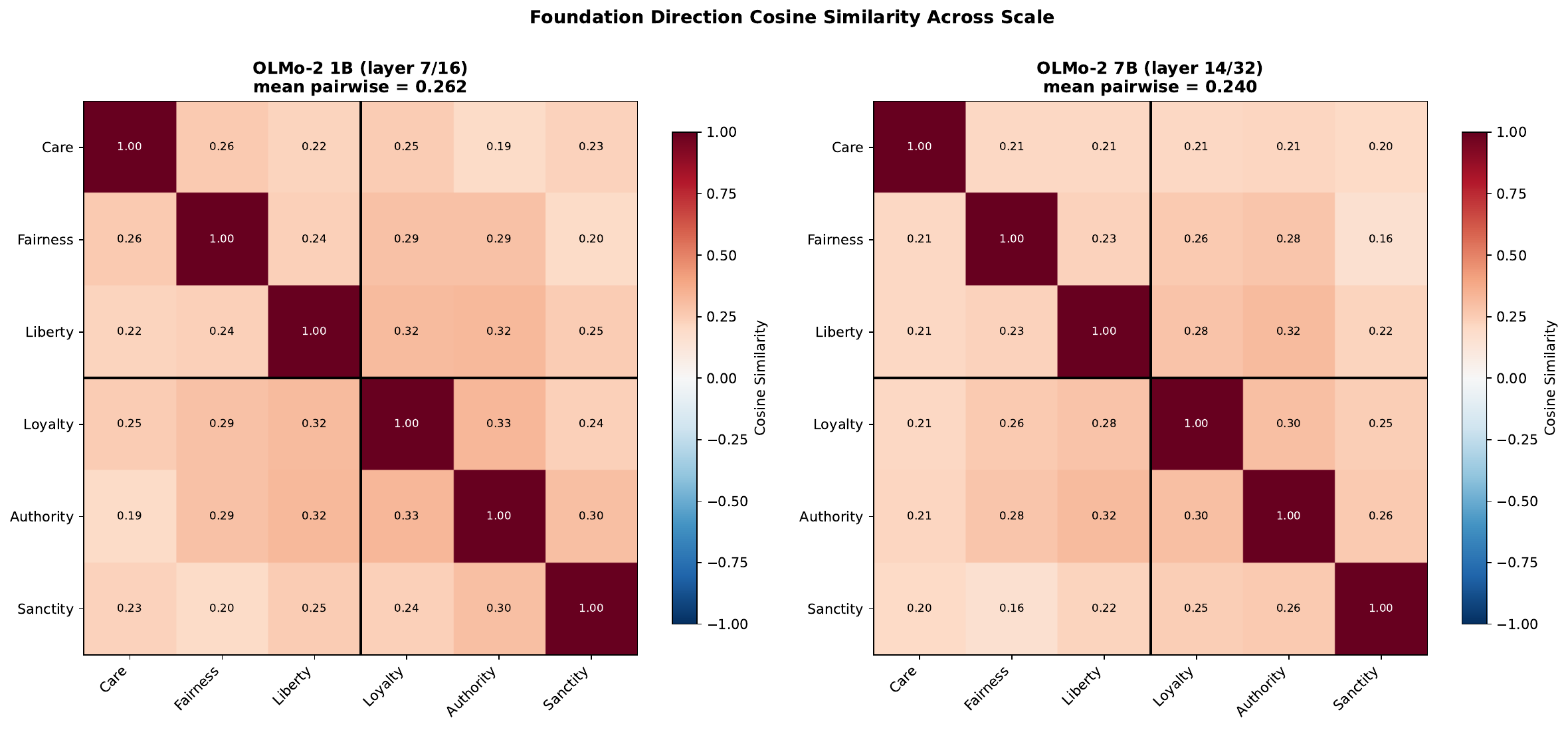}
\caption{Foundation direction cosine similarity for OLMo-2 1B (layer 7/16) and 7B (layer 14/32), at matched relative depth. Mean off-diagonal cosine is 0.262 (1B) and 0.240 (7B) at these display layers; both sit in the integration range. Effective dimensionality is 5 for both (Figure \ref{fig:scale_overlay}).}
\label{fig:scale_heatmap}
\end{figure}

\begin{figure}[t]
\centering
\includegraphics[width=\linewidth]{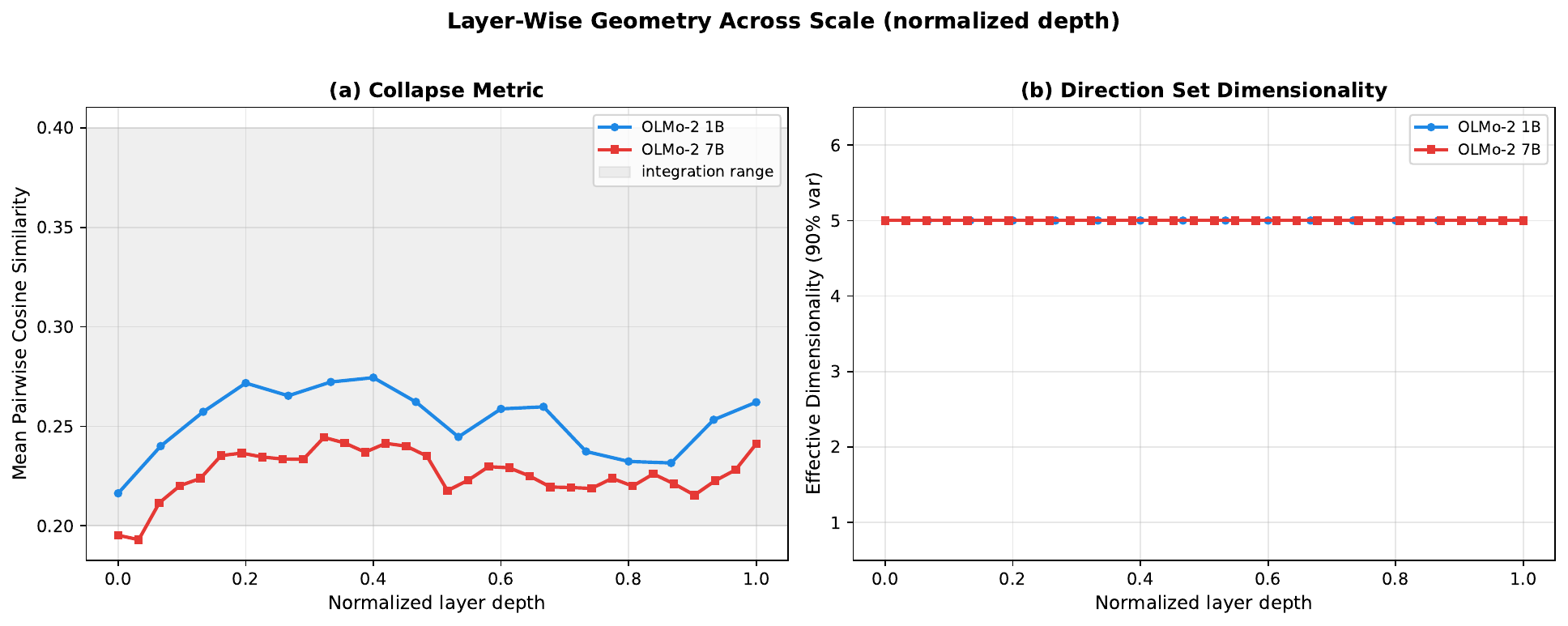}
\caption{Layer-wise geometry across scale on normalized depth. (a) Mean pairwise cosine for both models stays in the integration band. (b) Effective dimensionality is constant at 5 for both 1B and 7B.}
\label{fig:scale_overlay}
\end{figure}

\textbf{Direction stability.} Bootstrap resampling (200 iterations) places
110 of 192 foundation--layer combinations above the 0.8 stability
threshold, with the remainder in the 0.72--0.80 borderline band.
Sanctity/degradation is again the most stable foundation (29/32
layers), fairness/cheating the least (7/32). The per-layer cosine
range (0.72--0.85) is comparable to the 1B model; the larger model's
directions are no less determined.

\textbf{MFT is not recovered at 7B either.} Hierarchical clustering does
not produce the individualizing/binding split at any layer, and the
permutation test for MFT group structure is non-significant throughout
(exact \(p = 0.30\)--\(0.90\) across the 32 layers; floor \(0.10\)). The model's inter-framework geometry is not
MFT-aligned at either scale.

\textbf{Fragility.} Seed-averaged per-foundation fragility at 7B uses the
extended noise grid (\Cref{differential-fragility-across-frameworks}), which censors only 1 of 192 foundation--layer
cells at the cap. As at 1B, the per-foundation \(\sigma^*\) values have
wide, overlapping bootstrap CIs and no foundation is reliably most or
least robust (7B mean \(\sigma^*\): loyalty \(14.7\), authority \(13.5\),
care \(10.1\), sanctity \(10.0\), liberty \(8.4\), fairness \(6.9\), all with
overlapping CIs). The binding-vs-individualizing group difference is again
not significant (\(+4.2\), exact \(p = 0.20\)). What the 7B run does add is
a clean scale effect: the dense model is uniformly more robust at 7B
than at 1B (mean \(\sigma^*\) \({\approx}10.6\) vs.~\(5.0\)). The robust
fragility findings are therefore the cross-architecture dilution (\Cref{differential-fragility-across-frameworks})
and this dense scale effect, not any per-foundation ordering.

\begin{figure}[t]
\centering
\includegraphics[width=\linewidth]{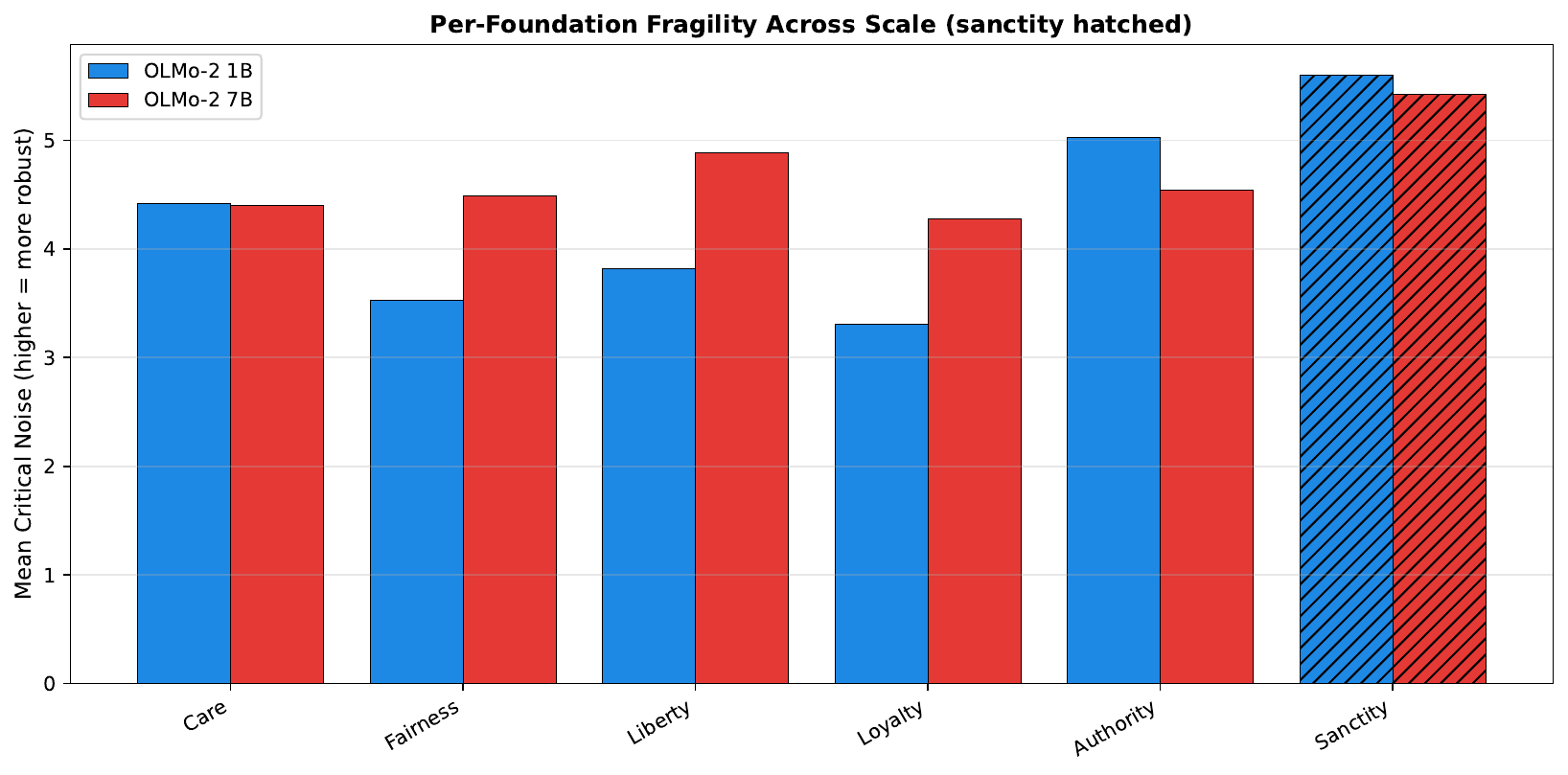}
\caption{Per-foundation mean critical noise $\sigma^*$ for OLMo-2 1B and 7B (dense), seed-averaged with bootstrap 95\% CIs. The 7B model is uniformly more robust (scale effect); within each model the per-foundation CIs overlap and no foundation is reliably most robust.}
\label{fig:scale_fragility}
\end{figure}

\subsection{The model's data-driven taxonomy does not recover MFT}\label{the-models-data-driven-taxonomy-does-not-recover-mft}

Sections 4.3 and 4.14 show that hierarchical clustering of the six
probe \emph{directions} does not recover MFT groups. We test the stronger
question directly: cluster the moral-positive \emph{activations} themselves,
without reference to foundation labels, and ask whether the discovered
groups align with the foundations.

K-means and spectral clustering (\(k\) selected by silhouette score,
\(k = 2\)--\(8\)) on OLMo-2 7B moral activations produce clusters that
barely align with MFT foundations. Adjusted mutual information between
the discovered clusters and the foundation labels peaks at \textbf{0.032}
(layer 11) and is near zero or slightly negative at most layers;
silhouette scores stay low (0.06--0.08), indicating weak cluster
structure. A clustering that recovered the foundations would give AMI
near 1.

\begin{figure}[t]
\centering
\includegraphics[width=\linewidth]{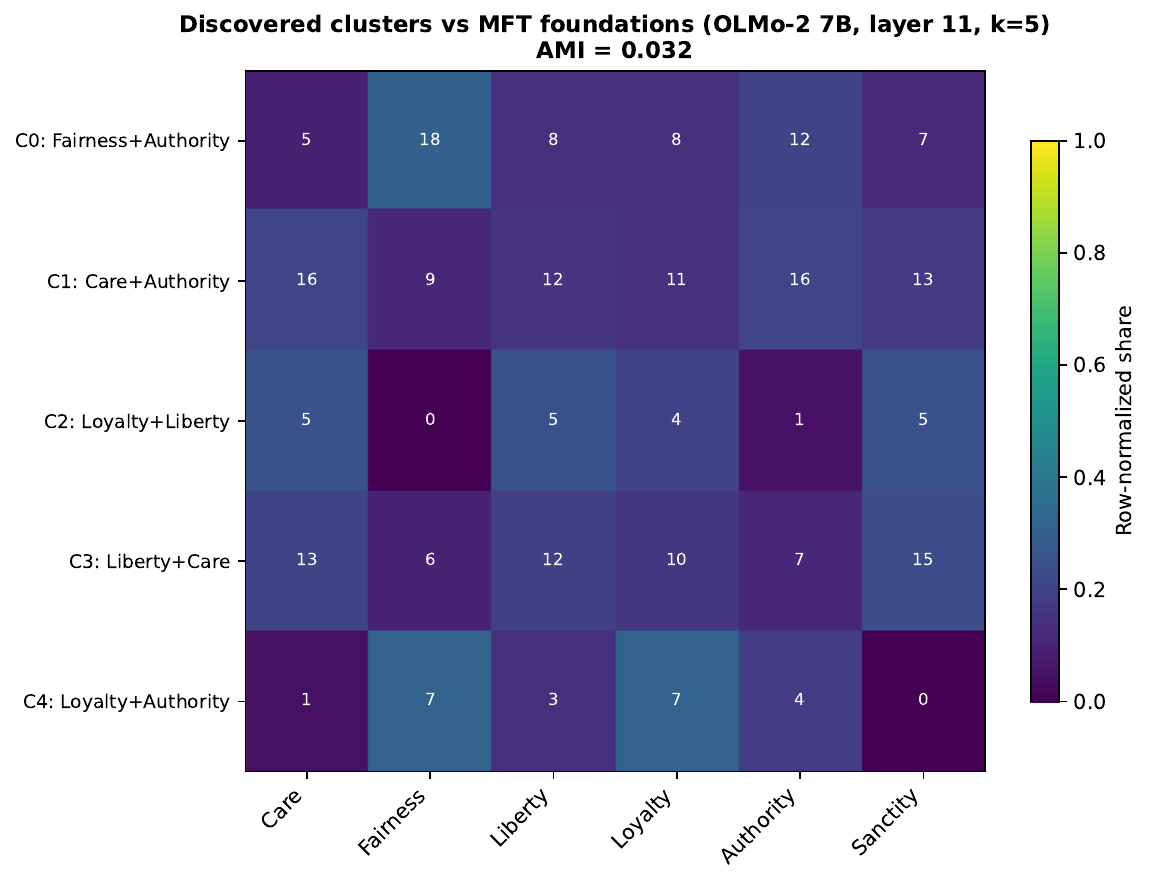}
\caption{Discovered clusters (rows) against MFT foundations (columns) at the most foundation-aligned layer (11) of OLMo-2 7B. Cells are pair counts; row-normalized shading. The clusters do not map onto foundations (AMI = 0.032).}
\label{fig:taxonomy_confusion}
\end{figure}

This is consistent with integration. Foundations are linearly
\emph{decodable}: supervised probes separate them with near-perfect
accuracy and span 5 effective dimensions. But they are not the
dominant axis of variation in the raw activations, so unsupervised
clustering recovers the shared moral-salience structure rather than
the foundation distinctions. The model represents the foundations as
distinct directions without organizing its moral activations around
them.

\subsection{External replication on the Moral Foundations Vignettes}\label{external-replication-on-the-moral-foundations-vignettes}

To test whether the integration geometry is a property of the model
rather than of our probing dataset, we recompute foundation directions
from the Moral Foundations Vignettes \citep{clifford2015mfv}, an
independently constructed and separately validated stimulus database.
We use the verbatim original English vignettes from the published norms
released by the authors (\citealp{clifford2015mfv};
\path|Full-Published-Norms.xlsx|, Table 1 respondent ratings),
selecting the five highest-loading items per foundation by the published
classification rate (the fraction of respondents who assigned a vignette
to its intended foundation); the care foundation draws its five from the
combined emotional- and physical-care pool. This yields a 30-item subset
(5 per foundation), none of which overlaps our probing dataset. Directions are estimated by
mean difference against a shared neutral baseline, the same estimator
applied to both datasets for a fair comparison.

The integration signature replicates on the genuine validated items.
The six MFV directions span \textbf{5 effective dimensions at every layer}
of OLMo-2 7B, identical to the directions derived from our dataset.
Mean pairwise cosine is higher for MFV (0.42--0.62 across layers, 0.62
at the display layer 8) than for the matched mean-difference directions
on our data (0.41 at the display layer), an expected small-sample
effect given five vignettes per foundation. Because
absolute cosine is inflated by the five-per-foundation sample, the
cross-dataset claim rests on the sign and structure of the geometry
rather than the cosine magnitude: both datasets show uniformly
positive mean cosine with variance concentrated on a shared leading
axis (the integration signature), spanning the same 5 effective
dimensions. Effective dimensionality by itself does not separate
integration from isolation (random directions also span \({\sim}5\)
dimensions); the positive mean cosine carries that distinction, and
it replicates in sign across the two datasets.

\begin{figure}[t]
\centering
\includegraphics[width=\linewidth]{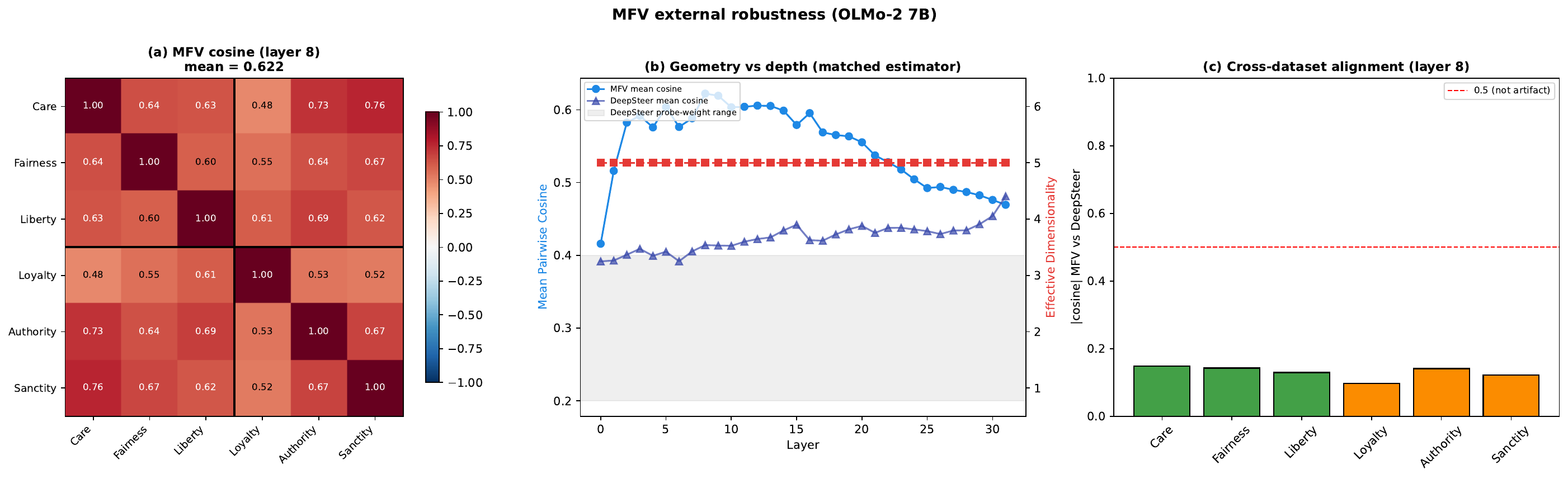}
\caption{MFV external replication on OLMo-2 7B. (a) Cosine similarity among the six MFV foundation directions. (b) Mean cosine and effective dimensionality vs depth for MFV and for matched mean-difference directions on our dataset. (c) Per-foundation alignment between MFV and our directions.}
\label{fig:external}
\end{figure}

\textbf{Cross-dataset direction alignment.} Foundation directions estimated
from the two independent datasets are positively aligned at every
foundation (\(|\cos| = 0.07\)--\(0.21\) per foundation at layer 12, matched
mean-difference estimator: care 0.19, fairness 0.14, liberty 0.21,
loyalty 0.17, authority 0.07, sanctity 0.20). Individualizing
foundations align marginally more strongly on average (mean 0.18) than
binding (mean 0.15), but the gap is small and there is no clean group
separation. The alignment is modest in magnitude, as expected from
30 verbatim vignettes versus a 240-pair probing set, but it is positive
throughout: that directions estimated from independently authored,
separately validated stimuli agree at all is evidence that the
foundation directions reflect the model's representations rather than
artifacts of any single dataset.

\section{Discussion}\label{discussion}

\subsection{Integration as the default geometric mode}\label{integration-as-the-default-geometric-mode}

The central finding, that foundation probe directions exhibit
integration rather than collapse or isolation, has a
straightforward interpretation: language models develop structured
moral representations from distributional statistics alone. The
training corpus does not label texts with MFT foundations, yet the
model learns representations in which moral foundations occupy
distinct directions that share a positive common component (mean
cosine \(\approx 0.22\) at peak separation, \(\approx 0.26\) at stable
mid-network layers). This is genuine multi-dimensional moral
structure, not a single ``moral salience'' detector. The shared component
is moral-specific relative to a matched non-moral concept battery built
identically to the foundations: that battery gives a mean cosine of
0.013 against the moral 0.26 (paired \(\Delta = 0.223\), CI
\([0.202, 0.244]\), excluding 0; \Cref{framework-geometry-integration-not-collapse}), and the estimator's
\([1+(k-1)\bar{c}]/k\) PC1 identity is confirmed to \(<0.01\) across all
three calibration constructions. The one control we do not run, and so
the one claim we do not make, is whether this axis is specifically moral
rather than generic affective salience (\S5.6).

The fact that effective dimensionality is 5 (near-maximal for 6
directions) throughout the network rules out the possibility that
the model has learned a single ``this text is moral'' feature and
then adds minor perturbations per foundation. The foundation
directions are geometrically distinct objects that happen to share
a common moral-salience component.

This holds across scale, architecture, and dataset. Effective
dimensionality is 5 at every layer of OLMo-2 1B, OLMo-2 7B, and
OLMoE-1B-7B, and on the independently constructed Moral Foundations
Vignettes (\S4.14, \S4.16). Integration is the default geometric mode
for moral representations in these models (all from Ai2's OLMo family;
\S5.6), not a property of one scale, architecture, or probing dataset.

\textbf{Dataset sensitivity.} The mean cosine similarity is sensitive to
neutral-pair quality: neutrals that inadvertently carry moral content
inflate the shared moral-salience component, producing higher cosine
values. The quality-gated dataset used here (\Cref{probing-dataset}) minimizes this
inflation, and we interpret the \({\approx}0.22\)--\(0.27\) range as a lower
bound on the true integration signal. All qualitative conclusions
(integration signature, effective dimensionality of 5, absence of MFT
dendrogram structure) are robust to dataset construction choices,
but the quantitative sensitivity highlights the importance of
neutral-pair quality for geometric analyses. The replication on the
Moral Foundations Vignettes (\S4.16), where directions estimated from
independently authored stimuli reproduce the 5-dimensional integration
geometry, shows that the qualitative signature does not depend on our
dataset.

\textbf{No evidence of MFT group structure.} The inter-framework
structure that emerges shows no alignment with MFT's predicted
individualizing/binding distinction: hierarchical clustering does
not recover this partition at any layer, and the permutation test
is non-significant throughout. This is an underpowered null, not a
demonstrated absence: the test enumerates only 10 distinct splits, so its
smallest achievable \(p\) is \(0.10\) and it cannot reject at
\(\alpha = 0.05\) (observed minimum \(0.40\); \Cref{dendrogram-structure-does-not-recover-mft-groups}), we
state no minimum detectable within/between gap, and we did not run a
positive control confirming the test fires on planted group structure,
so a small individualizing/binding effect cannot be excluded. Instead,
the most consistent clustering feature is a care--sanctity pairing that
crosses MFT groups. Both care and sanctity involve protection (of persons from
harm, of sacred things from degradation), sharing distributional
signatures that the model detects. The care--sanctity
pairing is itself robust to dataset construction choices: it persists
across different neutral-pair generation methods and quality
thresholds, arguing against a dataset artifact explanation. The
structure the model does form is thus empirically grounded in corpus
statistics and is not detectably aligned with the a priori
individualizing/binding grouping from moral psychology on this dataset,
consistent with genuine structure learning (the model discovers which
moral concepts are distributionally related) rather than surface
keyword matching.

A data-driven test sharpens this point. Clustering the moral
activations themselves, without foundation labels, recovers the
foundations only weakly (adjusted mutual information 0.03 at best;
\S4.15). Even the model's own unsupervised grouping of moral content
is not MFT-structured, which is the expected consequence of
integration: the foundations are linearly decodable but are not the
dominant axis of variation, so unsupervised methods find the shared
moral-salience structure instead.

\subsection{Per-foundation fragility is not separable}\label{per-foundation-fragility-is-not-separable}

Within an architecture, the moral foundations are not reliably
distinguishable by fragility. With 10 noise seeds and the cap-at-max
convention (\Cref{differential-fragility-across-frameworks}), per-foundation \(\sigma^*\) values carry wide,
overlapping bootstrap CIs in all three models, and no foundation is
reliably most or least robust. Sanctity sits mid-pack everywhere
(dense 1B \(5.50\), MoE \(2.33\), dense 7B \(10.0\), fourth of six), and the
binding-vs-individualizing group difference is not significant in any
model (dense 1B \(p = 0.40\), MoE \(p = 0.70\), dense 7B \(p = 0.20\)) and
reverses sign between dense and MoE. Per-foundation \(\sigma^*\) is
sensitive to single-draw noise, so we make no per-foundation or
MFT-group fragility claim.

The fragility result that does hold is between architectures, not
between foundations: every foundation is more fragile in MoE than in
dense (a \({\sim}2.3\times\) per-foundation gap, the same direction as the
pooled \(4.2\times\) of \citet{reblitzrichardson2026dilution}), and dense
robustness rises with scale from 1B to 7B. Output dilution suppresses
moral encoding uniformly across foundations rather than singling any
one out.

\textbf{What critical noise does and does not measure.} Three of our raw
\(\sigma^*\) comparisons turned out to be driven by the same thing once
controlled: activation-scale differences between the conditions being
compared. The per-foundation ordering (\Cref{differential-fragility-across-frameworks}), the cross-architecture
gap above, and the complexity gradient (\S4.11) each shrink or vanish
under RMS normalization, and the companion dense-model study reports the
same for its layer-depth gradient
(\citealp{reblitzrichardson2026fragility}, \Cref{layer-wise-geometric-development}). These are one
confound, not three. They sharpen what the metric is for: raw \(\sigma^*\)
measures \emph{practical perturbation sensitivity}, the absolute noise a
representation tolerates, which is real and useful, but it does not
measure encoding robustness independent of scale when the compared
conditions differ in activation scale, as they do across layers,
registers (declarative vs.~narrative), architectures, and probe
complexity. RMS-normalized \(\sigma^*\) is the right tool for those
cross-condition claims; raw \(\sigma^*\) is valid only where scale is
controlled by construction (within-layer, matched-stimulus contrasts).
Establishing that partition is a methodological contribution in its own
right.

\subsection{Framework geometry stabilizes before accuracy saturates}\label{framework-geometry-stabilizes-before-accuracy-saturates}

The trajectory analysis shows a temporal dissociation: framework
geometry (mean cosine similarity between foundation directions)
enters its integration regime within the first few thousand steps
(peaking near step 5000), while probing accuracy continues climbing
through step 10000 and beyond. This extends the ``accuracy saturates
but fragility resolves'' finding of
\citet{reblitzrichardson2026fragility} to a third metric: the
\emph{structure} of moral representations is laid down before the
\emph{strength} of those representations finishes developing.

This is consistent with a two-phase account of moral representation
learning. In the first phase (steps 0--5000) the model discovers the
geometric layout: the foundation directions move from near-orthogonal
into the integration band as a shared moral-salience component forms.
In the second phase (steps 5000+) the layout is not frozen but
gradually refined: the mean cosine declines slowly (0.34 $\to$ 0.29,
\({\sim}13\%\)) while accuracy is flat, so the foundations keep
differentiating from one another after they are individually
well-separated. The structure is set early and sharpened slowly,
not established once and held fixed.

Effective dimensionality is 5 from step 0 onward, indicating that
even random initialization produces probe directions that span 5
dimensions. This is expected: random unit vectors in
\(\mathbb{R}^{2048}\) are nearly orthogonal with high probability.
The informative signal is not dimensionality but cosine similarity,
which moves from \(\approx 0\) (random) into the integration band
(\({\sim}0.3\)) over the first few thousand steps and then drifts down.

\subsection{Partial compositionality of moral dilemmas}\label{partial-compositionality-of-moral-dilemmas}

When two moral foundations conflict in a dilemma scenario, the
model's representation is partially compositional: the dilemma probe
direction has more overlap with the 2D subspace of its own component
foundations (mean peak membership \(S = 0.118\)) than with the
mismatched-pair baseline of foundation pairs it shares no component
with (\(0.044\), a \({\sim}2.7\times\) margin that holds at every layer),
yet \({\sim}88\)\% of the dilemma direction lies outside its component
subspace.

This partial compositionality has two complementary
interpretations. First, the model recognizes moral dilemmas as
involving their component foundations (the matched-over-mismatched
margin is not an artifact), as confirmed by the shared-component
structure: dilemma pairs that share a foundation have higher cosine
similarity (\(\Delta = 0.076\) at layer 13, exact permutation
\(p = 0.0001\)). Second, the model represents something beyond the sum
of parts: the \({\sim}90\)\% residual captures conflict-specific
features (tension, trade-off framing, or contextual modulation)
that single-foundation probes do not isolate.

The near-balanced component loading (\(\bar{\alpha} = 0.486\)) is
notable. If dilemma representations were dominated by one
foundation (e.g., always prioritizing care over authority), we
would expect strongly asymmetric projections. Instead, both
conflicting foundations contribute roughly equally to the
within-subspace component, consistent with the model encoding the
\emph{conflict itself} rather than a pre-resolved moral judgment. This
is an ``understanding before choice'' mode of representation: the
model maintains both competing moral claims in tension rather than
collapsing to a resolution, suggesting that moral comprehension
(the capacity to represent the structure of ethical disagreement)
may precede and be separable from moral judgment. This pattern is
scale-stable: OLMo-2 7B reproduces the partial-compositionality
profile (9.1\% subspace membership, 0.95 residual, 0.46 component
balance; \S4.9), so the conflict-as-tension representation is not an
artifact of the smaller model's capacity.

A raw complexity--fragility ordering (single-foundation
\(\sigma^* = 5.02\) \(>\) pooled dilemma \(3.81\) \(>\) per-type dilemma
\(3.55\)) might suggest that representational complexity trades off
against robustness. It does not survive scale normalization: with
noise scaled to each layer's activation RMS the single-foundation and
pooled-dilemma values converge (both at the grid maximum) and the
per-type value barely separates (\S4.11). The raw ordering reflects
register-dependent activation scale, not differential encoding
robustness, so we make no complexity--robustness claim. The
compositionality evidence in \S4.9--4.10 (subspace membership,
shared-component geometry, balanced loading) is geometric, not
fragility-based, and is unaffected.

\subsection{Register sensitivity: directions transfer, thresholds do not}\label{register-sensitivity-directions-transfer-thresholds-do-not}

Foundation-specific probes trained on declarative minimal pairs do
not fully generalize to narrative dilemma text \emph{as classifiers}. In
the dilemma verification experiment, authority and loyalty probes
showed near-chance transfer (Youden's \(J < 0.2\)), while care and
fairness probes transferred well. This asymmetry is not a
model-capacity issue: testing on OLMo-2 7B (32 layers, 4096 hidden
dim) yielded comparable transfer failure (54.0\% vs.~61.3\% for
the 1B model).

\textbf{Directions vs.~thresholds.} The probe engineering analysis
(\S4.13) resolves this concern by separating two components of
cross-register transfer: the \emph{direction} (probe weight vector) and
the \emph{threshold} (bias term). When evaluated by pair accuracy (the
fraction of pairs where the direction projects the moral text higher
than the neutral text, requiring no threshold), both probe-weight
and mean-difference directions achieve \(>90\)\% mean pair accuracy on
narrative dilemma text (probe-weight \(>95\)\% for all foundations;
mean-difference drops to \({\sim}91\)\% for authority/subversion).
The Youden's \(J\) failure is therefore a threshold
miscalibration effect: the absolute projection scale shifts between
registers, invalidating the fixed decision boundary learned on
declarative text. The directional structure itself, the
subspace in which moral content is encoded, transfers robustly.

This distinction matters for the geometric findings. Cosine
similarity, effective dimensionality, and dendrogram clustering
depend only on direction vectors, not on classification thresholds.
Since the directions transfer across registers, the geometric
findings are not register-bound.

\textbf{Threshold vs.~direction transfer.} The register sensitivity is
specific to threshold transfer, not direction transfer. Both
individualizing and binding directions transfer across registers with
high pair accuracy (\(>90\)\%); what shifts across register is the
decision threshold, not the direction itself. Authority and loyalty
show the largest threshold gaps (Appendix B.2), so their register
sensitivity reflects a calibration shift rather than a change in the
underlying moral direction.

\textbf{Implications for the geometric findings.} The 21-direction
dendrogram analysis (\S4.9) gives direct evidence that register
features drive part of the representation geometry: projecting all
directions into the 5D moral subspace dissolves the categorical
foundation/dilemma separation, confirming that the separation is
carried by extra-moral (register) features. This is consistent with
the threshold miscalibration account: foundation and dilemma
directions occupy the same moral subspace (their projections
overlap), but differ in extra-moral dimensions that carry register
information and shift the activation scale.

\subsection{Limitations}\label{limitations}

\textbf{Small probing dataset.} The 32 training pairs per foundation
are sufficient for classification (near-perfect accuracy) but
limit the precision of direction estimation. Bootstrap analysis
(\Cref{direction-stability-under-bootstrap}) confirms that directions at layers 0--5 are borderline
unstable, and all geometric claims are qualified to the
bootstrap-stable core, layers 6--15 (the lone exception being
authority at layer 8, 0.792).
A larger dataset would tighten direction estimates, but the current
dataset is deliberately minimal to demonstrate that structured
geometry is recoverable even from small samples.

\textbf{Permutation test power.} With 6 foundations divided into two
groups of 3, the permutation space contains 20 partitions that form
10 mirror pairs (swapping the two groups leaves the statistic
unchanged), so the smallest achievable \(p\) is \(2/20 = 0.10\), reached
only if the MFT split is the single most extreme split. The test
therefore cannot reject at \(\alpha = 0.05\) on any data; the observed
minimum is \(0.40\) (\Cref{dendrogram-structure-does-not-recover-mft-groups}). We did not compute a minimum detectable
within/between gap or run a positive control verifying that the test
fires on planted group structure. Our MFT-group result is therefore ``no
evidence of individualizing/binding organization,'' not a demonstrated
absence: a small group effect cannot be excluded. The absence of
MFT-aligned dendrograms at any layer is consistent with this null but
does not turn it into a positive claim.

\textbf{One model family.} The three configurations span scale (1B and 7B
dense) and architecture (dense and MoE), and the integration signature
and MFT-mismatch hold across all of them and on the independently
constructed MFV stimuli (\S4.14--4.16). The models are still all from
Ai2's OLMo family, trained on comparable corpora, so generalization to
models trained on substantively different data mixtures remains open.
The architectural and scale comparisons are well-controlled because
the models share training data, at the cost of corpus diversity.

\textbf{Linear probes.} The entire analysis assumes that moral
foundations are encoded as linear directions. If some foundations
are encoded nonlinearly (e.g., as curved manifolds or distributed
across multiple directions), the cosine similarity analysis
would understate the true geometric richness. The near-perfect
accuracy of linear probes suggests that linear decoding captures
the dominant signal, but does not rule out additional nonlinear
structure.

\textbf{Affective vs.~moral salience.} The shared component that marks
integration is moral-specific relative to a matched non-moral battery
spanning affective, syntactic, stylistic, and topical concepts
(sentiment, register, grammaticality, tense, number, topic; \Cref{framework-geometry-integration-not-collapse}), each
of which decodes at 1.00 peak accuracy while giving a near-zero pairwise
cosine (0.013 vs.~the moral 0.26). That battery does not isolate
whether the shared moral axis is specifically \emph{moral} or a generic
\emph{evaluative/affective} salience common to emotionally charged text,
moral statements included. The cheapest discriminating control is a
matched-twin non-moral valence/affective battery (positive vs.~negative
affect against matched neutrals, built exactly as the foundation probes
are); we did not run it here. We therefore claim moral-specificity
relative to a matched non-moral battery and leave affective-vs-moral as
the open residual.

\textbf{Causal status.} Probe directions are read off the representation:
on their own they identify \emph{where} foundation information is readable,
not whether that information is \emph{used} during generation. Our 7B
foundation directions provide preliminary, uncontrolled causal checks:
ablating a foundation's direction selectively perturbs that
foundation's continuations, and injecting it produces a dose-response
shift (Appendix \ref{app:causal}). These checks carry no
random-direction or channel-matched null, so they do not yet separate
foundation-specific action from the generic effect of projecting out
(or amplifying) any stable direction, and the largest steering effects
occur at off-distribution injection strengths (\(\alpha = 20\)). Full
causal localization is left to future work; the geometry reported here
is descriptive of the representation, and we treat this causal evidence
as preliminary rather than as establishing that the directions are
functionally implicated.

\section{Conclusion}\label{conclusion}

Language models develop structured moral representations that go
beyond mere moral detection. By training independent linear probes
for each of six Moral Foundations Theory foundations and analyzing
the geometry of their weight vectors, we find that OLMo-2 1B and
OLMoE-1B-7B exhibit the \emph{integration} signature: foundation
directions are distinct (effective dimensionality \(= 5\), which rules
out collapse but, computed on the mean-centered direction matrix, is
at its ceiling for six directions and does not by itself distinguish
integration from isolation) yet share a common component, shown by a
uniformly positive mean pairwise cosine (\(\approx 0.22\)--\(0.27\)). This
positive cosine, not the effective dimensionality, is what
distinguishes integration from isolation; it is \({\sim}20\times\) a
matched non-moral concept battery built identically to the foundations
(0.26 vs.~0.013; paired \(\Delta = 0.223\), CI \([0.202, 0.244]\),
excluding 0; \Cref{framework-geometry-integration-not-collapse}), so the shared component is moral-specific relative
to that battery, with generic affective salience the one residual we
flag (\S5.6). (The leading
principal component captures \({\sim}0.38\) of the directions' variance,
vs.~\({\sim}0.18\) for random directions, but for near-equicorrelated
directions this is algebraically a re-expression of the mean cosine,
not independent evidence.) This
geometric structure is consistent across the two architectures
tested (dense vs.~MoE), emerges early in pre-training, and
stabilizes before probing accuracy saturates. However, we find no
evidence that the inter-framework structure aligns with MFT's predicted
individualizing/binding grouping; the group-structure test is
underpowered (smallest achievable \(p = 0.10\), so it cannot reject at
\(\alpha = 0.05\)), and a small effect is not excluded. The structure the model does form is grounded in corpus
statistics rather than the a priori human-theoretical grouping.

Framework-specific fragility testing shows that the output dilution
effect is foundation-uniform: every foundation is more fragile in MoE
than in dense (a \({\sim}2.3\times\) per-foundation gap), with no
foundation reliably most or least robust within an architecture and no
significant binding/individualizing group difference.

Extending this geometric lens to moral dilemmas (scenarios
where two foundations conflict) shows partial compositionality:
dilemma representations overlap their component-foundation subspace
\({\sim}2.7\times\) more than a mismatched-pair baseline (peak
membership 0.118 vs.~0.044, holding at every layer; shared-component
permutation \(p = 0.0001\)), with near-balanced loading across both
conflicting foundations. The remaining \({\sim}88\)\% residual indicates
that dilemma representations encode conflict-specific structure beyond
their component foundations. (A raw complexity--fragility gradient
across probe types does not survive scale normalization and is not used
as evidence; \S4.11.) This compositionality pattern is preserved across
architectures (dense and MoE) and from 1B to 7B.

The probe direction geometry methodology introduced here is
general. Any set of related concepts that can be isolated by
binary linear probes can be analyzed for inter-concept structure
via the same cosine similarity and clustering techniques. We
anticipate applications to other taxonomies of human values, to
political orientation, and to the internal organization of safety
training in instruction-tuned models.

\begin{ack}
This work made extensive use of Anthropic's Claude (the Claude Code agent on
Opus~4.6, 4.7, 4.8 and 5.5, and Fable~5 and 5.1) for code scaffolding, experimental scripts, and
prose drafting. The author retains responsibility for experimental design, all
scientific claims, and final wording.
\end{ack}

\bibliography{references}

\appendix
\newpage
\begin{center}
  \rule{0.5\linewidth}{0.4pt}\\[0.6em]
  {\Large\bfseries Appendices}\\[0.25em]
  {\small Supplementary material.}\\[0.4em]
  \rule{0.5\linewidth}{0.4pt}
\end{center}
\vspace{0.5em}

\section{Per-Foundation Probe Accuracy Tables}\label{per-foundation-probe-accuracy-tables}

\label{app:accuracy}

\subsection{OLMo-2 1B}\label{a.1-olmo-2-1b}

\begin{table}[h]
\centering
\caption{Per-foundation probe accuracy across layers, OLMo-2 1B. Each cell shows test accuracy (16 test examples per foundation). All foundations exceed 0.6 at every layer.}
\label{tab:accuracy_olmo}
\small
\begin{tabular}{r cccccc}
\toprule
Layer & Care & Fairness & Liberty & Loyalty & Authority & Sanctity \\
\midrule
0  & 1.000 & 1.000 & 1.000 & 0.938 & 1.000 & 0.938 \\
1  & 0.875 & 0.938 & 0.938 & 0.938 & 1.000 & 0.938 \\
2  & 0.938 & 0.938 & 0.938 & 0.938 & 1.000 & 0.938 \\
3  & 0.938 & 0.938 & 0.938 & 0.938 & 1.000 & 0.938 \\
4  & 1.000 & 0.938 & 1.000 & 0.938 & 1.000 & 0.938 \\
5  & 1.000 & 0.938 & 1.000 & 1.000 & 1.000 & 1.000 \\
6  & 1.000 & 0.938 & 1.000 & 1.000 & 1.000 & 0.938 \\
7  & 1.000 & 1.000 & 1.000 & 1.000 & 1.000 & 0.938 \\
8  & 1.000 & 1.000 & 1.000 & 1.000 & 1.000 & 1.000 \\
9  & 1.000 & 0.938 & 1.000 & 1.000 & 1.000 & 0.938 \\
10 & 1.000 & 0.938 & 1.000 & 1.000 & 1.000 & 0.938 \\
11 & 1.000 & 0.938 & 1.000 & 0.938 & 1.000 & 0.938 \\
12 & 1.000 & 0.938 & 1.000 & 0.938 & 1.000 & 1.000 \\
13 & 1.000 & 0.938 & 1.000 & 1.000 & 1.000 & 1.000 \\
14 & 1.000 & 0.938 & 1.000 & 1.000 & 1.000 & 0.938 \\
15 & 1.000 & 0.938 & 1.000 & 0.938 & 1.000 & 0.938 \\
\bottomrule
\end{tabular}
\end{table}

\subsection{OLMoE-1B-7B}\label{a.2-olmoe-1b-7b}

\begin{table}[h]
\centering
\caption{Per-foundation probe accuracy across layers, OLMoE-1B-7B.}
\label{tab:accuracy_olmoe}
\small
\begin{tabular}{r cccccc}
\toprule
Layer & Care & Fairness & Liberty & Loyalty & Authority & Sanctity \\
\midrule
0  & 0.938 & 0.875 & 0.938 & 0.938 & 0.875 & 0.938 \\
1  & 0.938 & 0.875 & 0.812 & 0.812 & 0.812 & 0.812 \\
2  & 0.938 & 0.875 & 0.875 & 0.812 & 0.875 & 0.812 \\
3  & 0.938 & 0.938 & 0.938 & 0.875 & 0.875 & 0.875 \\
4  & 0.938 & 0.938 & 0.938 & 1.000 & 1.000 & 1.000 \\
5  & 0.938 & 1.000 & 1.000 & 1.000 & 1.000 & 1.000 \\
6  & 0.938 & 0.938 & 1.000 & 1.000 & 1.000 & 1.000 \\
7  & 1.000 & 1.000 & 1.000 & 1.000 & 1.000 & 1.000 \\
8  & 1.000 & 1.000 & 1.000 & 1.000 & 1.000 & 1.000 \\
9  & 1.000 & 1.000 & 1.000 & 1.000 & 1.000 & 1.000 \\
10 & 1.000 & 0.938 & 1.000 & 1.000 & 1.000 & 1.000 \\
11 & 1.000 & 1.000 & 1.000 & 1.000 & 1.000 & 1.000 \\
12 & 1.000 & 0.938 & 1.000 & 1.000 & 1.000 & 1.000 \\
13 & 1.000 & 1.000 & 1.000 & 1.000 & 1.000 & 1.000 \\
14 & 1.000 & 0.938 & 1.000 & 1.000 & 1.000 & 1.000 \\
15 & 1.000 & 0.938 & 1.000 & 1.000 & 1.000 & 1.000 \\
\bottomrule
\end{tabular}
\end{table}

\section{Bootstrap Direction Stability Tables}\label{bootstrap-direction-stability-tables}

\label{app:bootstrap}

\begin{table}[h]
\centering
\caption{Bootstrap direction stability (mean cosine similarity with full-data direction, 200 resamples). Values marked with $^*$ fall below the 0.8 stability threshold.}
\label{tab:bootstrap}
\small
\begin{tabular}{r cccccc}
\toprule
Layer & Care & Fairness & Liberty & Loyalty & Authority & Sanctity \\
\midrule
0  & 0.740$^*$ & 0.741$^*$ & 0.768$^*$ & 0.761$^*$ & 0.780$^*$ & 0.793$^*$ \\
1  & 0.752$^*$ & 0.765$^*$ & 0.775$^*$ & 0.763$^*$ & 0.780$^*$ & 0.789$^*$ \\
2  & 0.774$^*$ & 0.779$^*$ & 0.796$^*$ & 0.785$^*$ & 0.790$^*$ & 0.789$^*$ \\
3  & 0.767$^*$ & 0.788$^*$ & 0.791$^*$ & 0.784$^*$ & 0.808 & 0.811 \\
4  & 0.789$^*$ & 0.803 & 0.817 & 0.796$^*$ & 0.812 & 0.819 \\
5  & 0.799$^*$ & 0.814 & 0.812 & 0.818 & 0.809 & 0.830 \\
6  & 0.807 & 0.816 & 0.830 & 0.819 & 0.809 & 0.825 \\
7  & 0.817 & 0.811 & 0.813 & 0.818 & 0.830 & 0.842 \\
8  & 0.821 & 0.822 & 0.822 & 0.815 & 0.792$^*$ & 0.807 \\
9  & 0.821 & 0.816 & 0.832 & 0.814 & 0.820 & 0.803 \\
10 & 0.823 & 0.809 & 0.831 & 0.821 & 0.817 & 0.836 \\
11 & 0.817 & 0.802 & 0.830 & 0.837 & 0.833 & 0.814 \\
12 & 0.828 & 0.801 & 0.831 & 0.843 & 0.828 & 0.820 \\
13 & 0.825 & 0.811 & 0.821 & 0.817 & 0.827 & 0.838 \\
14 & 0.821 & 0.823 & 0.824 & 0.842 & 0.826 & 0.827 \\
15 & 0.816 & 0.826 & 0.803 & 0.847 & 0.838 & 0.852 \\
\bottomrule
\end{tabular}
\end{table}

Layers 0--2 have widespread instability (\(< 0.8\)): all six
foundations are unstable. Layers 3--5 are mixed (2--6 of 6
unstable). Layers 6--15 are mostly stable, with one exception
(authority at layer 8, 0.792). Sanctity/degradation is the most
stable foundation (13/16 layers above threshold); care/harm is
the least stable (10/16).

The stable core (layers 6--15) gives the most reliable geometric
measurements. The headline finding (integration signature, effective
dimensionality = 5) is confirmed within this range.

\section{Cosine Similarity Matrices}\label{cosine-similarity-matrices}

\label{app:cosine}

Full pairwise cosine similarity matrices between the six foundation
probe directions at selected layers of OLMo-2 1B.

\begin{table}[h]
\centering
\caption{Cosine similarity matrix at layer 0 (peak separation), OLMo-2 1B. Mean off-diagonal = 0.216.}
\label{tab:cosine_l0}
\small
\begin{tabular}{r cccccc}
\toprule
     & Care & Fair & Lib & Loy & Auth & Sanc \\
\midrule
Care & 1.000 & 0.148 & 0.189 & 0.203 & 0.164 & 0.201 \\
Fair & 0.148 & 1.000 & 0.232 & 0.179 & 0.224 & 0.143 \\
Lib  & 0.189 & 0.232 & 1.000 & 0.273 & 0.333 & 0.224 \\
Loy  & 0.203 & 0.179 & 0.273 & 1.000 & 0.264 & 0.219 \\
Auth & 0.164 & 0.224 & 0.333 & 0.264 & 1.000 & 0.249 \\
Sanc & 0.201 & 0.143 & 0.224 & 0.219 & 0.249 & 1.000 \\
\bottomrule
\end{tabular}
\end{table}

\begin{table}[h]
\centering
\caption{Cosine similarity matrix at layer 7, OLMo-2 1B. Mean off-diagonal = 0.262.}
\label{tab:cosine_l7}
\small
\begin{tabular}{r cccccc}
\toprule
     & Care & Fair & Lib & Loy & Auth & Sanc \\
\midrule
Care & 1.000 & 0.264 & 0.221 & 0.250 & 0.188 & 0.229 \\
Fair & 0.264 & 1.000 & 0.236 & 0.294 & 0.292 & 0.195 \\
Lib  & 0.221 & 0.236 & 1.000 & 0.317 & 0.324 & 0.252 \\
Loy  & 0.250 & 0.294 & 0.317 & 1.000 & 0.329 & 0.242 \\
Auth & 0.188 & 0.292 & 0.324 & 0.329 & 1.000 & 0.298 \\
Sanc & 0.229 & 0.195 & 0.252 & 0.242 & 0.298 & 1.000 \\
\bottomrule
\end{tabular}
\end{table}

\begin{table}[h]
\centering
\caption{Cosine similarity matrix at layer 15, OLMo-2 1B. Mean off-diagonal = 0.262.}
\label{tab:cosine_l15}
\small
\begin{tabular}{r cccccc}
\toprule
     & Care & Fair & Lib & Loy & Auth & Sanc \\
\midrule
Care & 1.000 & 0.246 & 0.200 & 0.226 & 0.168 & 0.195 \\
Fair & 0.246 & 1.000 & 0.220 & 0.347 & 0.278 & 0.138 \\
Lib  & 0.200 & 0.220 & 1.000 & 0.352 & 0.393 & 0.237 \\
Loy  & 0.226 & 0.347 & 0.352 & 1.000 & 0.415 & 0.220 \\
Auth & 0.168 & 0.278 & 0.393 & 0.415 & 1.000 & 0.295 \\
Sanc & 0.195 & 0.138 & 0.237 & 0.220 & 0.295 & 1.000 \\
\bottomrule
\end{tabular}
\end{table}

At layer 0, the highest pairwise cosine similarity is
liberty--authority (0.333), followed by liberty--loyalty (0.273).
At later layers, the liberty--authority and loyalty--authority pairs
consistently show the strongest similarity (0.393 and 0.415 at
layer 15). These pairings cross the MFT individualizing/binding
boundary, indicating that the model's inter-framework geometry is
organized along empirical distributional axes rather than the
theoretical MFT grouping.

\section{Permutation Tests for MFT Group Structure}\label{permutation-tests-for-mft-group-structure}

\label{app:permutation}

We test whether the six foundation probe directions cluster into the
MFT-predicted individualizing (care, fairness, liberty) and binding
(loyalty, authority, sanctity) groups. The test statistic is the
difference between mean within-group cosine similarity and mean
between-group cosine similarity. With six foundations split into two
groups of three, there are only \(\binom{6}{3} = 20\) group
assignments, so we enumerate the null distribution exactly rather than
resampling; the \(p\)-value is the fraction of the 20 assignments whose
statistic is \(\geq\) the observed value, with a statistic within
\(10^{-12}\) of the observed one counted as a tie. The 20 assignments
form 10 mirror pairs (either group can be listed first) with equal
statistics, so \(p\) is a multiple of \(0.10\) and its smallest attainable
value is \(0.10\).

\begin{table}[h]
\centering
\caption{Exact permutation test for individualizing/binding group structure across layers, OLMo-2 1B (enumeration over all 20 assignments; attainable floor $p = 0.10$). No layer reaches significance.}
\label{tab:permutation}
\small
\begin{tabular}{r cc}
\toprule
Layer & Observed statistic & $p$-value \\
\midrule
0  &  0.001 & 0.40 \\
1  & $-$0.014 & 0.80 \\
2  & $-$0.003 & 0.70 \\
3  & $-$0.004 & 0.80 \\
4  & $-$0.004 & 0.80 \\
5  &  0.003 & 0.50 \\
6  & $-$0.002 & 0.50 \\
7  &  0.005 & 0.40 \\
8  &  0.001 & 0.50 \\
9  & $-$0.003 & 0.60 \\
10 & $-$0.003 & 0.60 \\
11 & $-$0.014 & 0.80 \\
12 & $-$0.004 & 0.70 \\
13 &  0.003 & 0.40 \\
14 &  0.000 & 0.40 \\
15 &  0.007 & 0.40 \\
\bottomrule
\end{tabular}
\end{table}

The test does not reach significance at any layer (minimum \(p = 0.40\)).
Its attainable floor is \(p = 0.10\), reached only when the MFT split is
the single most extreme of the 10 distinct splits, so the test cannot
reject at \(\alpha = 0.05\) on any data and a small group effect cannot
be excluded. At no layer does the MFT split rank above fourth of the
ten. The observed statistics are near zero and frequently
negative (within-group similarity \(<\) between-group similarity),
confirming that the model's inter-framework geometry is not
organized along the MFT individualizing/binding axis. This is
consistent with the dendrogram analysis (\Cref{dendrogram-structure-does-not-recover-mft-groups}), which shows
cross-MFT pairings (care--sanctity, liberty--authority) rather
than the predicted group structure.

\section{Reproducibility}\label{reproducibility}

\label{app:reproducibility}

\subsection{Hardware and software}\label{e.1-hardware-and-software}

All experiments ran on a single MacBook Pro with Apple M4 Pro
(24 GB unified memory) using the MPS backend. Software versions:
Python 3.13, PyTorch 2.7, Transformers 4.49.

\subsection{Runtime}\label{e.2-runtime}

\begin{table}[h]
\centering
\small
\begin{tabular}{llr}
\toprule
Experiment & Model & Wall time \\
\midrule
1--3 (probing + geometry + bootstrap) & OLMo-2 1B & $\sim$5 min \\
5 (dense vs.\ MoE geometry) & OLMoE-1B-7B & $\sim$20 min \\
6 (geometric trajectory) & OLMo-2 1B (37 ckpts) & $\sim$45 min \\
7 (framework fragility) & OLMo-2 + OLMoE & $\sim$25 min \\
\bottomrule
\end{tabular}
\end{table}

\subsection{Reproducibility notes}\label{e.3-reproducibility-notes}

\textbf{Probe training.} Linear probes are \texttt{nn.Linear(2048,\ 1)} trained
with BCE loss and Adam (lr = \(10^{-2}\)) for 50 epochs. No weight
decay, no learning rate schedule. Random seed is not fixed across
runs; bootstrap analysis (Appendix \ref{app:bootstrap}) quantifies
direction stability under resampling.

\textbf{Probing dataset.} The 240-pair minimal-pair dataset (40 per MFT
foundation) is deterministic and version-controlled. Dataset
generation uses Claude Sonnet 4.6 with automated validation gates
(embedding similarity, keyword scan, LLM-as-judge filtering); the
exact dataset is included in the code repository.

\textbf{Activation collection.} Forward passes use \path|torch.no_grad()|.
Activations are mean-pooled across the sequence dimension at each
layer. For OLMoE, activations are collected after the expert
combination step (post-routing), not from individual experts.

\textbf{Code availability.} All experiment scripts, the probing dataset,
and figure generation code are available at
\url{https://github.com/deepsteer/deepsteer}.

\section{Causal Validation (Preliminary)}\label{causal-validation-preliminary}

\label{app:causal}

The geometry reported in the main text is read off the
representation: it shows where foundation information is decodable.
Whether the model \emph{uses} that information during generation is a
separate, causal question. Here we report preliminary, uncontrolled
causal checks on the OLMo-2 7B foundation directions used throughout
this paper. They carry no random-direction or channel-matched null, so
they are consistent with the directions being functionally implicated
but do not by themselves rule out the generic effect of intervening on
any stable direction. A full causal localization is left to future work.

\textbf{Direction ablation.} Projecting a foundation's direction out of the
residual stream at a target layer degrades that foundation's
continuations more than the other foundations'. Specificity, the gap
between the on-target and off-target mean effect over 48 prompts and
the six foundations, is negative at every layer tested and deepens
with depth, from \(-0.12\) at layer 4 to \(-0.32\) at layer 14. Removing a
foundation's direction selectively harms that foundation, and the
effect is strongest where the directions are most stable. These
numbers are on OLMo-2 7B (\path|allenai/OLMo-2-1124-7B|), layers 4--14, over
48 prompts (\path|outputs/probe_engineering_7B/direction_ablation_mean_diff.json|).

\textbf{Steering injection.} Adding \(\alpha\) times a foundation's direction
to the residual stream produces a dose-response. Mean specificity
rises monotonically with the injection strength, from \(+0.08\) at
\(\alpha = 1\) to \(+0.16\), \(+0.61\), \(+1.85\), and \(+3.59\) at
\(\alpha = 2, 5, 10, 20\). The same direction that decodes a foundation
also steers generation toward it, in proportion to the dose.

These results are descriptive and uncontrolled rather than a full
causal localization, which we leave to future work. Without a
random-direction or channel-matched specificity control they do not
separate foundation-specific action from generic intervention effects,
so we read them as consistent with the foundation directions in \Cref{results}
corresponding to features the model acts on during generation, not as
establishing it.

\section{Dilemma subspace membership: matched vs.~mismatched baseline}\label{dilemma-subspace-membership-matched-vs.-mismatched-baseline}

\label{app:dilemma_membership}

Per-layer mean subspace membership of the 15 dilemma directions on
OLMo-2 1B, averaged over pairs. \textbf{Matched} is membership in the 2D
span of each dilemma's own two component foundation directions;
\textbf{mismatched} is the mean membership in the 2D spans of all foundation
pairs that share no component with the dilemma (the correct null, since
it absorbs the shared moral-salience component that the random-vector
null, \({\sim}0.001\), does not). The matched value exceeds the mismatched
baseline at every layer.

\begin{table}[h]
\centering
\caption{Per-layer matched vs. mismatched dilemma subspace membership, OLMo-2 1B (mean over 15 dilemma pairs). Matched directions are seed-averaged probe-weight directions; mismatched is averaged over the foundation pairs sharing no component.}
\label{tab:dilemma_membership}
\small
\begin{tabular}{r ccc}
\toprule
Layer & Matched & Mismatched & Gap \\
\midrule
0 & 0.0664 & 0.0313 & +0.0351 \\
1 & 0.0723 & 0.0334 & +0.0389 \\
2 & 0.0861 & 0.0450 & +0.0411 \\
3 & 0.0939 & 0.0525 & +0.0414 \\
4 & 0.1007 & 0.0507 & +0.0500 \\
5 & 0.1057 & 0.0494 & +0.0563 \\
6 & 0.1058 & 0.0480 & +0.0579 \\
7 & 0.0968 & 0.0433 & +0.0535 \\
8 & 0.0944 & 0.0392 & +0.0552 \\
9 & 0.0905 & 0.0380 & +0.0524 \\
10 & 0.0831 & 0.0370 & +0.0461 \\
11 & 0.0913 & 0.0338 & +0.0574 \\
12 & 0.0907 & 0.0295 & +0.0612 \\
13 & 0.0912 & 0.0296 & +0.0617 \\
14 & 0.0939 & 0.0314 & +0.0624 \\
15 & 0.0991 & 0.0330 & +0.0662 \\
\bottomrule
\end{tabular}
\end{table}

Cross-layer means: matched 0.091, mismatched 0.039 (\({\sim}2.3\times\);
paired-bootstrap gap 0.052, CI \([0.037, 0.069]\), excluding 0);
per-pair-peak means: matched 0.118, mismatched 0.044 (\({\sim}2.7\times\);
gap 0.074, CI \([0.053, 0.100]\), excluding 0, but a max-over-layers
extremum and biased upward, so the cross-layer-mean gap is the unbiased
figure). Both bootstraps resample the 15 dilemmas (\(n = 10^4\)).
The same matched-over-mismatched margin replicates on OLMo-2 7B
(peak 0.090 vs.~0.032) and OLMoE-1B-7B (peak 0.118 vs.~0.045).

\end{document}